%% file: Paper.tex
\RequirePackage{fix-cm}
\documentclass[smallcondensed]{svjour3}     
\smartqed  
\usepackage{graphicx}
\usepackage{cite}
\usepackage{amsmath,amssymb,amsfonts}
\usepackage{algorithm}
\usepackage{makecell}
\usepackage{textcomp}
\usepackage{array} 
\usepackage{longtable}
\usepackage{booktabs}
\usepackage{float}
\usepackage{color,xcolor}
\usepackage{algpseudocode}
\usepackage{booktabs}
\usepackage{multirow}
\begin{document}

\title{A Hierarchical Conditional Random Field-based Attention Mechanism Approach for Gastric Histopathology Image Classification}

\titlerunning{Yixin Li et al.}        

\author{Yixin Li\and Xinran Wu \and Chen Li\and Changhao Sun 
\and Md Rahaman\and Haoyuan Chen\and Yudong Yao \and Xiaoyan Li\and Yong Zhang \and Tao Jiang 
}


\institute{Yixin Li \at
              Microscopic Image and Medical Image Analysis Group, 
              College of Medicine and Biological Information Engineering, 
              Northeastern University, China \\
              \email{20185414@stu.neu.edu.cn
}           
           \and
           Xinran Wu (co-frist author) \at
              Microscopic Image and Medical Image Analysis Group, 
              College of Medicine and Biological Information Engineering, 
              Northeastern University, China
              \and
              Chen Li (corresponding author) \at
              Microscopic Image and Medical Image Analysis Group, 
              College of Medicine and Biological Information Engineering, 
              Northeastern University, China\\
              \email{lichen201096@hotmail.com}
             \and
              Changhao Sun \at
              Microscopic Image and Medical Image Analysis Group, 
              College of Medicine and Biological Information Engineering, 
              Northeastern University, China; 
              Shenyang Institute of Automation, Chinese Academy of Sciences, Shenyang, China
              \and
              Md Rahaman \at 
              Microscopic Image and Medical Image Analysis Group, 
              College of Medicine and Biological Information Engineering, 
              Northeastern University, China 
              \and
              Haoyuan Chen \at 
              Microscopic Image and Medical Image Analysis Group, 
              College of Medicine and Biological Information Engineering, 
              Northeastern University, China 
              \and
              Yudong Yao \at 
              Department of Electrical and Computer Engineering, 
              Stevens Institute of Technology, Hoboken, NJ 07030, USA
              \and 
              Xiaoyan Li \at 
              China Medical University, Liaoning Cancer Hospital and Institute, Shenyang
              \and 
              Yong Zhang\at
              China Medical University, Liaoning Cancer Hospital and Institute, Shenyang
              \and 
              Tao Jiang \at Control Engineering College, 
              Chengdu University of Information Technology, 
              Chengdu People's Republic of China 
}

\date{Received: date / Accepted: date}

\maketitle

\begin{abstract}
In the \emph{Gastric Histopathology Image Classification} (GHIC) tasks, which are usually weakly supervised learning missions, there is inevitably redundant information in the images. Therefore, designing networks that can focus on distinguishing features has become a popular research topic. In this paper, to accomplish the tasks of GHIC superiorly and assist pathologists in clinical diagnosis, an intelligent \emph{Hierarchical Conditional Random Field based Attention Mechanism} (HCRF-AM) model is proposed.
The HCRF-AM model consists of an \emph{Attention Mechanism} (AM) module and an \emph{Image Classification} (IC) module. In the AM module, an HCRF model is built to extract attention regions. In the IC module, a \emph{Convolutional Neural Network} (CNN) model is trained with the attention regions selected, and then an algorithm called Classification Probability-based Ensemble Learning is applied to obtain the image-level results from the patch-level output of the CNN. In the experiment, a classification specificity of $96.67\%$ is achieved on a gastric histopathology dataset with 700 images. Our HCRF-AM model demonstrates high classification performance and shows its effectiveness and future potential in the GHIC field.

\end{abstract}
\input{introduction}

\input{relatedwork}

\input{method}

\input{Experiment}
\input{discussion}

\input{conclusion}

\input{ack}


\bibliographystyle{unsrt}
\bibliography{reference}

\end{document}

%% file: introduction.tex
\section{Introduction}
\label{sec:introduction}
According to the research of World Health Organisation (WHO), gastric cancer is one of the top five most frequently diagnosed malignant tumors worldwide~\cite{wild2014world}. It is considered to be the third leading cause of cancer deaths for its high incidence and fatality rate. Over 1,000,000 new cases and over 700,000 deaths per year are related to the disease~\cite{bray2018global}. The only possible cure in the early stage without metastasis is surgical removal. The median survival of gastric cancer rarely exceeds 12 months, and after the tumor metastasis, 5-years of survival is observed with a survival rate under 10\%~\cite{orditura2014treatment}. Therefore, early treatment can effectively reduce the possibility of death, and an accurate estimate of the patient's prognosis is demanded. Although endoscopic ultrasonography and Computerized Tomography (CT) are the primary methods for gastric cancer analysis, whereas histopathology images are considered as the gold standard for the diagnosis \cite{van2016gastric}. However, histopathology images are usually large with redundant information, which means histopathology analysis is a time-consuming specialized task and highly associated with pathologists' skill and experience \cite{elsheikh2013american}. Professional pathologists are often in short supply, and long hours of heavy work can lead to lower diagnostic quality. Thus, an intelligent diagnosis system has a significant effect on detecting and categorizing histopathology images automatically.

\begin{figure}[!htbp]
\centering
\centering{\includegraphics[width=0.9\columnwidth]{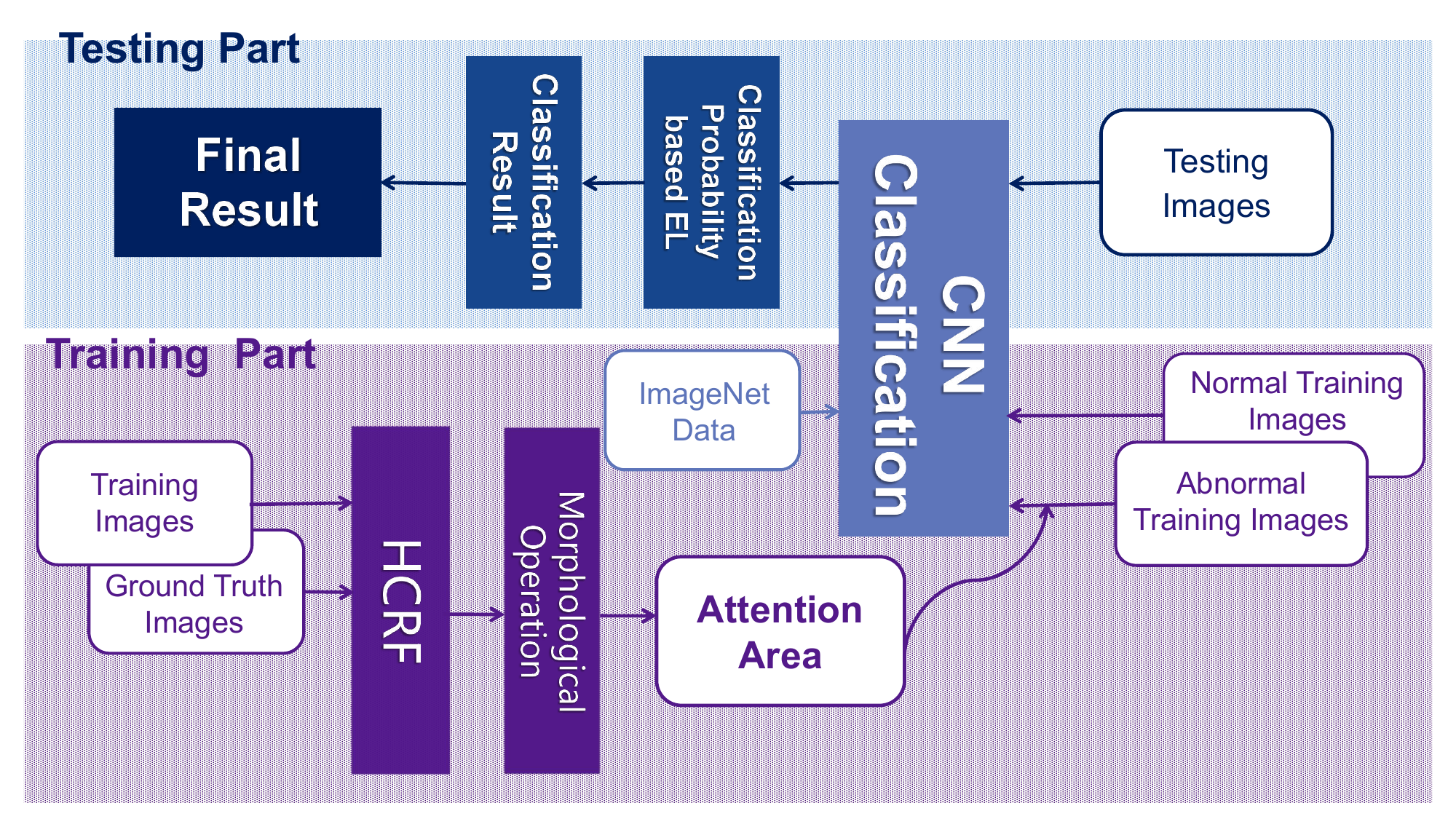}}
\caption{Flowchart of the proposed HCRF-AM model for GHIC.}
\label{fig:workflow}
\end{figure}

Recently, Deep Learning (DL) techniques have has received profound progress in a variety of computer vision tasks, consisting of diagnosis of gastric cancer, lung cancer and breast cancer. Especially, \emph{Gastric Histopathology Image Classification} (GHIC) task is usually a weakly supervised problem, which means that an image labeled as abnormal contains abnormal regions with cancer cells and normal regions without cancer cells existing in the surrounding area at the same time. However, previous networks are often incapable of focusing only on abnormal areas, which leads to noise regions and  surplus information, bringing negative influence on the final decision-making process and affecting the network performance~\cite{wang2019thorax}. Therefore, some advanced methods are proposed to incorporate visual \emph{Attention Mechanisms} (AMs) into \emph{Convolutional Neural Networks} (CNNs), which enables a DL model to adaptively concentrate on relevant areas of the images~\cite{li2019attention}. There are two prevalent ways to apply AM method. The first one is to change the block of the initial network. For example, the algorithm based on reinforcement learning, whose mechanism is to make the model pay more attention to some local details through the Reward function. Besides, some researchers accomplish their goals by applying the objective function and the corresponding optimization function. The second idea is combining the segmentation and object detection methods with the original network, which is utilized in our paper. Moreover, it is costly and time-consuming to acquire the full annotation like contours or bounding boxes of pathological samples. Hence, to guide the attention of CNNs for the GHIC tasks, we propose an intelligent \emph{Hierarchical Conditional Random Field based Attention Mechanism} (HCRF-AM) model including additional region level images. The HCRF-AM model includes the AM module (where the Hierarchical Conditional Random Field (HCRF) model \cite{sun2020gastric,sun2020hierarchical} is employed to obtain attention areas) and the \emph{Image Classification} (IC) module. The work-flow of our HCRF-AM model is illustrated in Fig
.~\ref{fig:workflow}. 

The main contributions of our work can be concluded in three parts: First, the AM module integrated into the network improves both the performance and interpretability of gastric cancer diagnosis. 
Second, we develop the HCRF model to obtain full annotations for weakly supervised classification tasks automatically. Thirdly, we apply a public gastric histopathology image database containing $700$ images. The extensive experiments on them prove the outstanding performance of our method.

This paper is divided into five sections. We review the existing methods related to automatic gastric cancer diagnosis, AMs, and the Conditional Random Field (CRF) in Sec.~\ref{sec:Related}. Afterward, we explain our proposed method in Sec.~\ref{sec:method}. Sec.~\ref{sec:Experiment} elaborates the experimental settings, implementation, results, and comparison. Sec.~\ref{sec:Discussion} compares our approach with existing GHIC researches. Sec.~\ref{sec:Conclusion} gives a summary of this paper and discusses the future work.

%% file: relatedwork.tex
\section{Related Works}
\label{sec:Related}
\subsection{Automatic Gastric Cancer Diagnosis}
Though different imaging techniques like gastroscopes~\cite{zhu2015lesion}, X-ray~\cite{ishihara2017detection}, and CT~\cite{li2018detection} are employed to diagnose gastric cancer, histopathological analysis of gastric cancer slides by pathologists is the gold standard that can be used to diagnose gastric cancer with confidence. Researchers have conducted a continuous and in‑depth discussion and there is numerous work on automatic GHIC tasks.

Here, we group Computer-Aided Diagnosis (CAD) methods of GHIC into two types: classical Machine Learning (ML) techniques and DL techniques. The classical ML methods extract some handcrafted features like color~\cite{li2020multi} and texture descriptors~\cite{korkmaz2017recognition}~\cite{korkmaz2018classification} and some classifiers like Support Vector Machine (SVM)~\cite{sharma2015appearance}~\cite{liu2018classification}, Random Forest (RF)~\cite{sharma2017comparative} and Adaboost algorithm~\cite{li2020multi} are adopted to make decision. Nevertheless, the classical ML methods mentioned above only take limited features on images into consideration, resulting in relatively low classification accuracy.

In recent years, numerous DL models have been proposed in the literature to diagnose gastric cancer with images obtained under the optical microscope. For instance, a pure supervised feedforward CNN model for gastric carcinoma Whole Slide Images (WSIs) classification task is introduced in~\cite{sharma2017deep}.
Compared with traditional image analysis methods, which require manual features to be calculated in advance, the performance of this method is obviously improved. The comparative experimental results reveal that DL methods compare favorably to traditional methods. The work in~\cite{liu2018gastric} creates a residual network for GHIC tasks, which has the characteristics of more complex structure, fewer parameters, and higher precision. A DL method employing a multi-instance learning algorithm is proposed in~\cite{wang2019rmdl} for WSI classification. 
This method offers a valid choice to utilize the interrelationships of different patches and take their various effects on the image-level label classification into consideration.
A CNN of DeepLab-v3 with the ResNet-50 architecture is applied for segmentation task in~\cite{song2020clinically}, and a private dataset containing 2123 Haematoxylin and Eosin (H\&E) stained WSIs the network is employed to train the network. A deep neural network that can learn multi-scale morphological patterns of histopathology images simultaneously is proposed in~\cite{kosaraju2020deep}. In order to classify WSIs into three categories, the work of~\cite{iizuka2020deep} contributes to reducing the number of parameters of standard Inception-v3 network by using a depth multiplier. The feature map of the Inception-v3   feeds in a Recurrent Neural Network with two Long Short-Term Memory layers.

Although existing methods based on DL models provide a significant performance boost in gastric histopathology image analysis, the existing methods still neglect that the images in weakly-supervised learning tasks contain large redundancy regions that are insignificant in the DL process which is the main challenge in computational pathology.

\subsection{Applications of Attention Mechanism}
The visual Attention Mechanism (AM) has the capacity to make a deep model adaptively concentrate on regions with distinctive features of an image. Therefore, it is an effective way to enhance the model performance in various computer vision tasks, including object detection~\cite{ba2014multiple},~\cite{li2020object}, image caption~\cite{xu2015show},~\cite{liu2020image} and action recognition~\cite{sharma2015action}. In a comprehensive review~\cite{chaudhari2019attentive}, some recent attention models are divided into different categories.  According to the difference of the structure, AM methods can be classified into Memory Networks, Transformer and Encoder-Decoder. Another commonly used classification method of AM is dividing them into soft attention and hard attention. The former pays more attention to regions or channels, and it is deterministic attention, which can be directly generated through the network after learning. The most important thing is that soft attention is differentiable. The attention that can be differentiated can be calculated through the neural network to calculate the gradient and forward propagation and backward feedback to learn the weight of attention. It calculates the weighted average of N input information and then puts it into the neural network for calculation, which is applied in our paper. The hard attention mechanism can select essential features from the input information, but the hard attention mechanism is more difficult to train. 

There are some previous works that employ AM method in the pathology analysis field. A prediction model for WSI analysis is proposed in~\cite{bentaieb2018predicting}, which integrates a recurrent AM. By selecting a limited sequence of positions adaptively, it is able to deal with discriminatory areas of the image.
An attention-based CNN is introduced in~\cite{li2019large}, where the attention prediction subnet is capable of emphasizing the salient areas for glaucoma detection. A DenseNet based guided soft attention network is developed in~\cite{yang2019guided} which aims at localizing regions of interest in breast cancer histopathology images and simultaneously using them to guide the classification network. In order to classify thorax diseases on chest radiographs, a Thorax-Net is constructed in~\cite{wang2019thorax}. The correlation among class labels is considered by applying the attention branch of the proposed network. The classification branch decides the locations of pathological abnormalities. Finally, by calculating the average of the two branches' outputs and binarizing, the final diagnosis is achieved. A CAD approach called HIENet is introduced in~\cite{sun2019computer} for endometrial disease diagnosis by integrating the AM block into CNNs. The Position Attention block of the HIENet is a self-AM, which is utilized to capture contextual relationships between different local regions of the input image. GHIC is essentially a weakly supervised learning task, where the location of key regions plays a critical role. Therefore, it is reasonable to combine the AMs in the classification of tissue-scale gastric histopathology images.

\subsection{Applications of Conditional Random Fields}
Conditional Random Fields (CRFs), an essential and popular type of ML method, are designed for building probabilistic models to explicitly describe the correlation of the pixels or the patches being predicted and label sequence data. The CRFs are attracting increasing attention in the field of ML for they allow achieving in various research fields, such as Natural Language Processing~\cite{zhang2017semi}, Information Mining~\cite{wicaksono2013toward}, Behavior Analysis~\cite{zhuowen2013human}, Image and Computer Vision~\cite{kruthiventi2015crowd}, and Biomedicine~\cite{liliana2017review}. In recent years, with the rapid development of DL, the CRF models are usually utilized as an essential pipeline within the deep neural network in order to refine the image segmentation results. Some research incorporates them into the network architecture, while others include them in the post-processing step. In~\cite{qu2019weakly}, in order to improve the accuracy and further improve the model, a dense CRF is embedded into the loss function of a CNN model. In~\cite{Zormpas2019Superpixel}, inspired by the way pathologists observe local tissue architecture, a hierarchical framework containing multi-resolution information is introduced. In the CRFs, the labels of target nodes are connected to corresponding superpixels prediction results. In~\cite{li2020automated}, a method based on a CNN is presented for prostate cancer diagnosis, where a CRF-based post-processing method is employed to refine the initial results. In~\cite{Dong2020GECNN}, a CNN containing group equivariant convolution module is presented. The output probability of the CNN model can build up the unary potential of the CRFs. By using the feature mapping of adjacent blocks, a pairwise loss function is designed to formulate the connection between two blocks.

In our previous work~\cite{kosov2018environmental}, an environmental microbial classification algorithm based on CRF and DL network is proposed. The segmentation accuracy of around 94.2\% is finally achieved. Besides, in order to classify cervical cancer images, a CRF model with multiple layers is introduced in~\cite{li2019cervical}. Finally, we manage to obtain an overall accuracy of 93\% on a cervical
histopathology image dataset. In~\cite{sun2020gastric}, we optimize our architecture and propose the HCRF model, which is employed to segment gastric cancer images for the first time. The results show overall better performance in comparison with other popular segmentation methods when the experiment was conducted on the same dataset. Furthermore, our research group combines the AM with the HCRF model and applies them in classification tasks, obtaining preliminary research results in~\cite{li2021intelligent}. To acquire more information, please refer to our previous review~\cite{li2020comprehensive}. The spatial dependencies on patches are usually neglected in previous GHIC tasks, and in the inference process, it only considers the appearance of individual patches. Thus, we describe an AM based on the HCRF framework in this paper, which has not been adopted in the domain of pathology.

%% file: method.tex
\section{Method}
\label{sec:method}
\begin{figure}[!htbp]
\centering
\centering{\includegraphics[width=1\columnwidth]{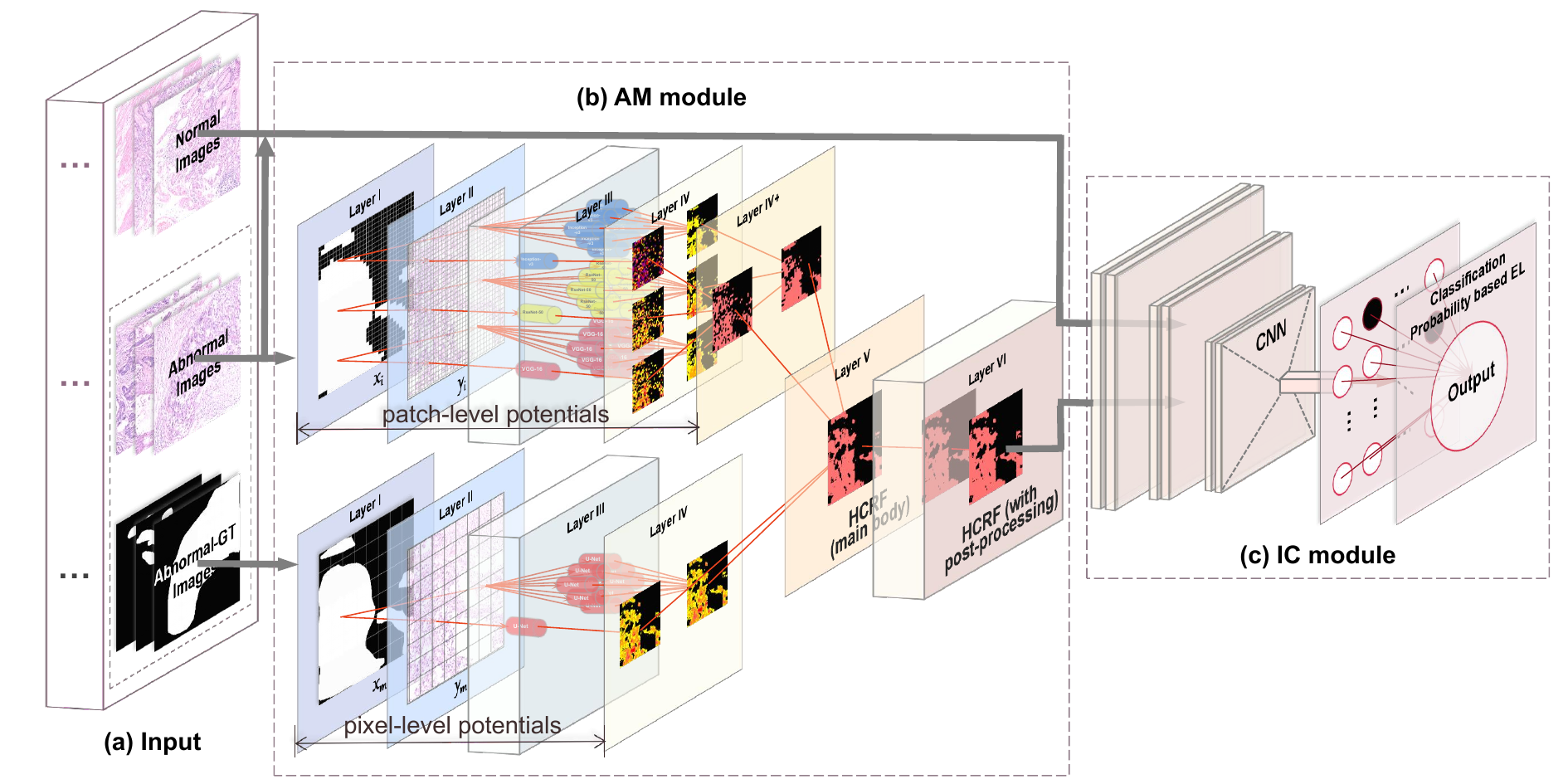}}
\caption{The architecture of the HCRF-AM model for gastric histopathological image classification
(a) The input of HCRF-AM model. 
(b) The AM module. 
(c) The IC module.}
\label{fig:structure}
\end{figure}

Various kinds of classifiers have been used in GHIC tasks, and CNN classifiers are proved to achieve better performance than some classical ML methods. 
However, the results obtained by training them directly are not so satisfying. Considering that fact, we develop the HCRF-AM model to refine the classification results further. Our proposed method consists of three main building blocks such as Attention Mechanism (AM) module, Image Classification (IC) module, and \emph{Classification Probability-based Ensemble Learning} (CPEL). Fig.~\ref{fig:structure} shows the architecture of our HCRF-AM model. We explain each building block in the following subsections.

\subsection{AM Module}
\label{ss:method:AM}
The AM module is integrated to assist the CNN classifier with extracting key characteristics of the abnormal images and reducing redundant information meanwhile. The HCRF, which is the improvement of CRF~\cite{lafferty2001conditional}, has excellent attention area detection performance because it can characterize the spatial relationship of images~\cite{li2020comprehensive}. The fundamental definition of CRFs will be introduced first. The detailed information of HCRF model including four kinds of potentials and their combination will be elaborated afterward.
\subsubsection{Basic Mathematics Knowledge of CRFs}
\label{ss:CRF:basic}
In \cite{lafferty2001conditional}, the definition and the generation process of CRF are first introduced. The essential mathematical symbol is listed in Table~\ref{tab:symbol}.

\begin{table}[h]
    \centering
    \renewcommand\arraystretch{1.1}
    \caption{The definition and brief introduction of the key mathematical symbol in CRF.}
    \scriptsize
    \begin{tabular}{|p{1.5cm}<{\raggedright}|p{4cm}<{\raggedright}|}
    \hline
        Mathematical symbol  & Description \\ \hline
        $ \textbf{Y}$/$\textbf{y}$ & The random variable of label sequence that can be observed \\ \hline
        $\textbf{X}$/$\textbf{x}$ & The random variable of the label sequence \\ \hline
        $V$ & The array of all sites/all nodes in the graph $G$ \\ \hline
        $E$ & The interactions among adjacent sites/the set of all edges in the graph $G$ \\ \hline
        $G$ & $\textit{G} = (\textit{V},\textit{E})$ is the undirected graph, where $\textbf{X}=(\textbf{X}_{v}) _{v\in\textit{V}}$, when $\textbf{X}$ is indexed by the vertices of G. \\ \hline
        $w \sim v$ & $w$ and $v$ are neighbours in $\textit{G}= (\textit{V},\textit{E})$ \\ \hline
        $\textit{p}(\textbf{X} | \textbf{Y})$ & Conditional distribution \\ \hline
        $\textit{C}$ & The subset of the vertices in the undirected graph $\textit{G}$, $C \subseteq V$ \\ \hline
        $\psi_{}(,)$  & The potential function over two items \\ \hline
    \end{tabular}
    \label{tab:symbol}
\end{table}
 The CRF model is built when $\textbf{X}_{v}$ and $\textbf{Y}$ follow the Markov properties related to the graph: $\textit{p}=(\textbf{X}_{v} | \textbf{Y}, \textbf{X}_{w}, w \neq v) = \textit{p}(\textbf{X}_{v} | \textbf{Y}, \textbf{X}_{w}, w \sim v)$.
These equations indicate the CRF model is an undirected graph where two disjoint sets, $\textbf{X}$ and $\textbf{Y}$, are divided from the nodes. According to the theory of the random fields in~\cite{Clifford-1990-MRF, Chen-2018-DSI, Zheng-2015-CRFRNN, Gupta-2006-CRF},  the joint distribution over the label sequence $\textbf{X}$ given $\textbf{Y}$ forms as Eq.~\ref{xz:equ:2}.
\begin{equation}
p(\textbf{X}|\textbf{Y})=\dfrac{1}{Z}\prod_{C}\psi_{C}(\textbf{X}_{C}, \textbf{Y}),
\label{xz:equ:2}
\end{equation}
where $Z=\sum_{\textbf{X}\textbf{Y}}P(\textbf{X}|\textbf{Y})$ denotes the 
normalization factor. 

\subsubsection{The Structure of the HCRF Model}
\label{sss:CRF:our:structure}
Unlike most of CRF models that constructed with only unary and binary potentials~\cite{Zheng-2015-CRFRNN, Chen-2018-DSI}, two types of high order potentials are developed in our HCRF. One is a patch-unary potential to describe target images characteristic, the other is a patch-binary potential to formulate the contextual relationship among neighboring tissue regions. The expression of HCRF is shown by Eq.~\ref{equ:1}.
\begin{equation}
\begin{aligned}
p(\textbf{X}|\textbf{Y})=&\dfrac{1}{Z}\prod_{i \in V}\varphi_{i}(x_{i};\textbf{Y};w_{V})
\prod_{(i,j) \in E}\psi_{(i,j)}(x_{i},x_{j};\textbf{Y};w_{E})\\
&\times\prod_{m \in V_{P}}\varphi_{m}({\mathrm x_{m}};\textbf{Y};w_{m};w_{V_{P}})\\
&\times\prod_{(m,n)\in E_{P}}\psi_{(m,n)}({\mathrm x_{m}},{\mathrm x_{n}};\textbf{Y};w_{(m,n)};w_{E_{P}}),
\label{equ:1}
\end{aligned}
\end{equation}
where 
\begin{equation}
\begin{aligned}
 Z=&\sum_{\textbf{X}\textbf{Y}}\prod_{i \in V}\varphi_{i}(x_{i};\textbf{Y})\prod_{(i,j) \in E}\psi_{(i,j)}(x_{i},x_{j};\textbf{Y})\\
&\times\prod_{m \in V_{P}}\varphi_{m}({\mathrm x_{m}};\textbf{Y})\prod_{(m,n)\in E_{P}}\psi_{(m,n)}({\mathrm x_{m}},{\mathrm x_{n}};\textbf{Y})
\label{equ:2}
\end{aligned}
\end{equation}
is the normalization factor;  
$V_{P}$ denotes one patch separated from an image; $E_{P}$ denotes the adjacent patches. 
The usual clique potential function contains two terms: the pixel-unary potential function $\varphi_{i}( x_{i},\textbf{Y})$, which is used to calculate the probability that a pixel node $i$ is labeled as $x_{i}\in\textbf{X}$, for a given observation vector 
$\textbf{Y}$~\cite{kosov2018environmental}; 
the pixel-binary potential function $\psi_{(i,j)}(x_{i},x_{j};\textbf{X})$ is used to describe the neighboring nodes $i$ and $j$ in the graph. 
The contextual connection between them is related  to the label of node $i$ and the label of its neighbour node $j$. 
Besides, patch-unary potential $\varphi_{m}({\mathrm x_{m}};\textbf{Y})$ and patch-binary potential $\psi_{(m,n)}({\mathrm x_{m}},{\mathrm x_{n}};\textbf{Y})$ are the newly developed high order potentials. 
The former is applied to calculate the possibility that a patch node $m$ is labeled as $\mathrm x_{m}$ for a 
given $\textbf{Y}$; 
the latter is utilized to formulate the neighboring nodes $m$ and $n$ in the patch. 
$w_{V}$, $w_{E}$, $w_{V_{P}}$ and $w_{E_{P}}$ are the weights of the four corresponding potentials. 
$w_{m}$ and $w_{(m,n)}$ are the weights of the $\varphi_{m}(\cdot;\textbf{Y})$ and 
$\psi_{(m,n)}(\cdot,\cdot;\textbf{Y})$, respectively. 
These weights are employed to figure out the largest posterior label 
$\tilde{\textbf{X}}=\textrm{arg}\,\textrm{max}_{\textbf{X}}\,p(\textbf{X}|\textbf{Y})$ 
and for performance optimization. 

The work-flow of the proposed HCRF  model can be concluded as follows: First, the popular segmentation network U-Net~\cite{ronneberger2015u} is trained to segment the images and build up the pixel-level potential at the same time. Second, for purpose of obtaining sufficient spatial information in patch-level, three pre-trained CNNs are fine-tuned , including Inception-V3 \cite{szegedy2016rethinking}, VGG-16 \cite{simonyan2014very}, and ResNet-50 \cite{he2016deep} networks to construct the patch-level potential. Thirdly, on the basis of the pixel- and patch-level potentials, our HCRF model is structured. In the AM module, a half of abnormal images and their Ground Truth (GT) images are applied to train the HCRF. Finally, in order to further refine the results, we perform some post-processing algorithm on the images obtained from the HCRF, such as closing operations and hole filling.

\subsubsection{Pixel-unary and Pixel-binary Potential}
\label{sss:CRF:our:pixel}
The pixel-unary potential $\varphi_{i}(x_{i};\textbf{Y};w_{V})$ in Eq.~\ref{equ:1} 
is related to the probability weights $w_{V}$ of a label $x_{i}$, taking a value 
$c\in\mathbb L$ given the observation data $Y$ by Eq.~\ref{equ:5}. 
\begin{equation}
\begin{aligned}
\varphi_{i}(x_{i};\textbf{Y};w_{V})\propto{\Big(p(x_{i}=c|f_{i}(Y))\Big)^{w_{V}}}, 
\end{aligned}
\label{equ:5}
\end{equation}
where the image content is characterized by site-wise feature vector $f_{i}(Y)$, which may 
be determined by all the observation data $Y$~\cite{Kumar-2006-DRF}. The probability maps $p(x_{i}=c|f_{i}(Y))$ at the last convolution layer of the U-Net serves as the feature maps, and the $256 \times 256 \times 2$-dimensional pixel-level feature $f_{i}(Y)$ obtains. 

The pixel-binary potential $\psi_{(i,j)}(x_{i},x_{j};\textbf{Y};w_{E})$ in 
Eq.~\ref{equ:1} describes the similarity of the pairwise adjacent sites $i$ and $j$ to 
take label $(x_{i},x_{j})=(c,c')$ given the data and weights, 
and it is defined as Eq.~\ref{equ:7}. 
\begin{equation}
\begin{aligned}
\psi_{(i,j)}(x_{i},x_{j};\textbf{Y};w_{E})\propto{}
\Big(p(x_{i}=c;x_{j}=c'| f_{i}(Y),f_{j}(Y))\Big)^{w_{E}}. 
\label{equ:7}
\end{aligned}
\end{equation}

\begin{figure}[!htbp]
\centering
\centering{\includegraphics[width=0.6\columnwidth]{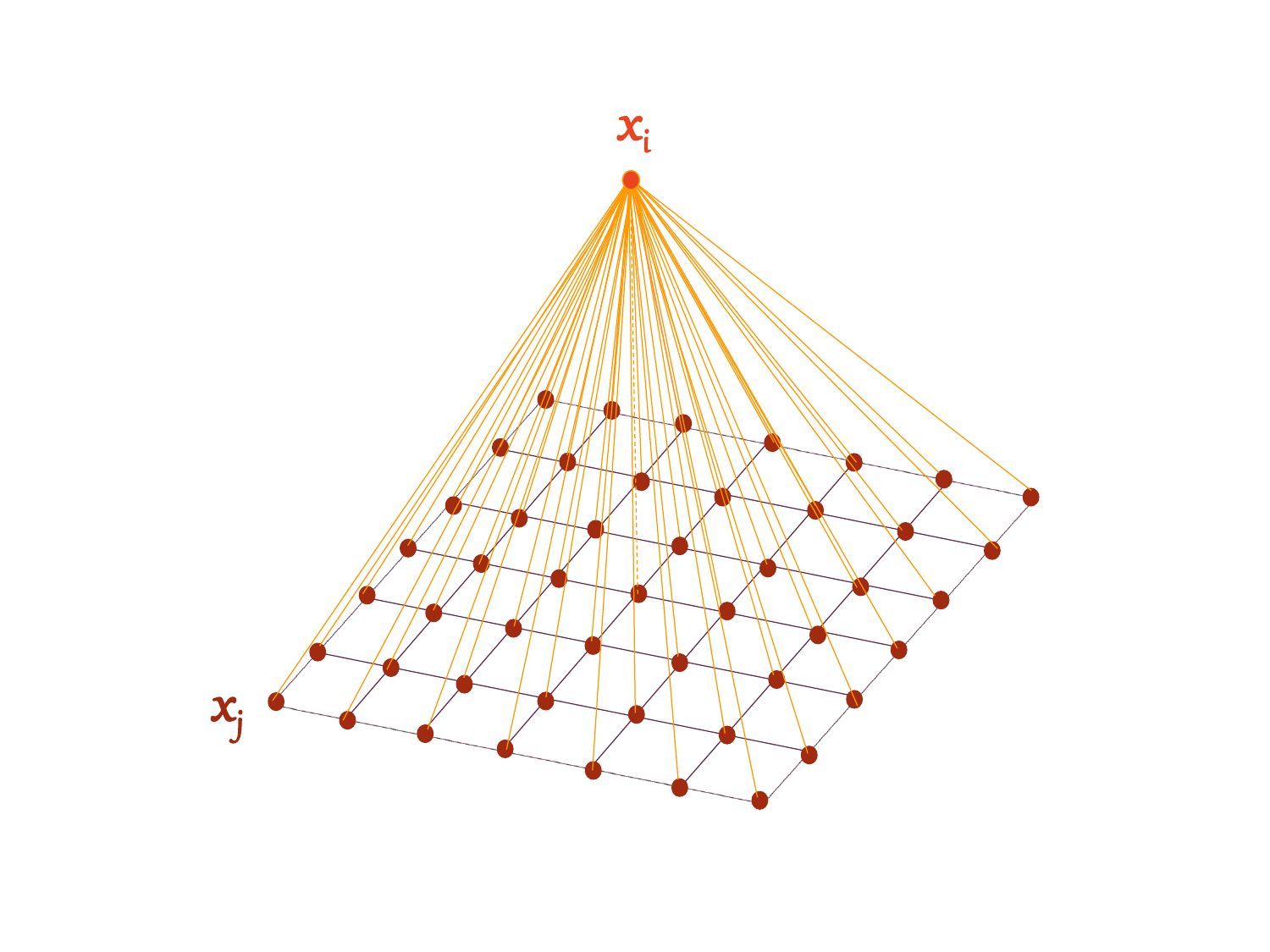}}
\caption{The special design ‘lattice’  of pixel-binary potential in AM module. Each variable has 48 neighbourhood. 
The probability of the pixel (central pixel in orange) is obtained by calculating the average of the unary probabilities of adjacent pixels.}\label{fig:small1}
\end{figure}

The design of the pixel-binary potential is illustrated in Fig.~\ref{fig:small1}.  The other procedures are the same as the pixel-unary potential calculation.

\subsubsection{Patch-unary and Patch-binary Potential}
\label{sss:CRF:our:patch}
In patch-level terms, $\alpha, \beta, \gamma$ represent VGG-16, Inception-V3 and ResNet-50 networks, which are selected to extract spatial information. In patch-unary potentials 
$\varphi_{m}({\mathrm x_{m}};\textbf{Y};w_{m};w_{V_{P}})$ of Eq.~\ref{equ:1}, 
label $\mathrm x_{m} =\{\mathrm x_{(m,\alpha)},\mathrm x_{(m,\beta)},\mathrm x_{(m,\gamma)}\}$ 
and $w_{m}=\{w_{(m,\alpha)},w_{(m,\beta)},w_{(m,\gamma)}\}$. 
$\varphi_{m}({\mathrm x_{m}};\textbf{Y};w_{m};w_{V_{P}})$ are related to the probability 
of labels $(w_{(m,\alpha)},w_{(m,\beta)},w_{(m,\gamma)})=(c,c,c)$ given the data $Y$ by 
Eq.~\ref{equ:8}. 
\begin{equation}
\begin{aligned}
&\varphi_{m}({\mathrm x_{m}};\textbf{Y};w_{m};w_{V_{P}})\propto{}
\Big((p(\mathrm x_{(m,\alpha)}=c|f_{(m,\alpha)}(Y)))^{w_{(m,\alpha)}}\\
&(p(\mathrm x_{(m,\beta)}=c|f_{(m,\beta)}(Y)))^{w_{(m,\beta)}}
(p(\mathrm x_{(m,\gamma)}=c|f_{(m,\gamma)}(Y)))^{w_{(m,\gamma)}}\Big)^{w_{V_{P}}},
\label{equ:8}
\end{aligned}
\end{equation}
where the characteristics in image data are transformed by site-wise feature vectors $f_{(m,\cdot)}(Y)$ ($\cdot$ represents $\alpha, \beta,\gamma$) that may be determined by 
all the input data $Y$. For $f_{(m,\cdot)}(Y)$, we use 1024-dimensional patch-level bottleneck features $F_{(m,\cdot)}$, obtained from pre-trained three models by ImageNet. In order to calculate the classification probability of each category, the last three layers~\cite{Kermany-2018-IMD} are retrained with our dataset. 

\begin{figure}[!htbp]
\centering
\centering{\includegraphics[width=0.6\columnwidth]{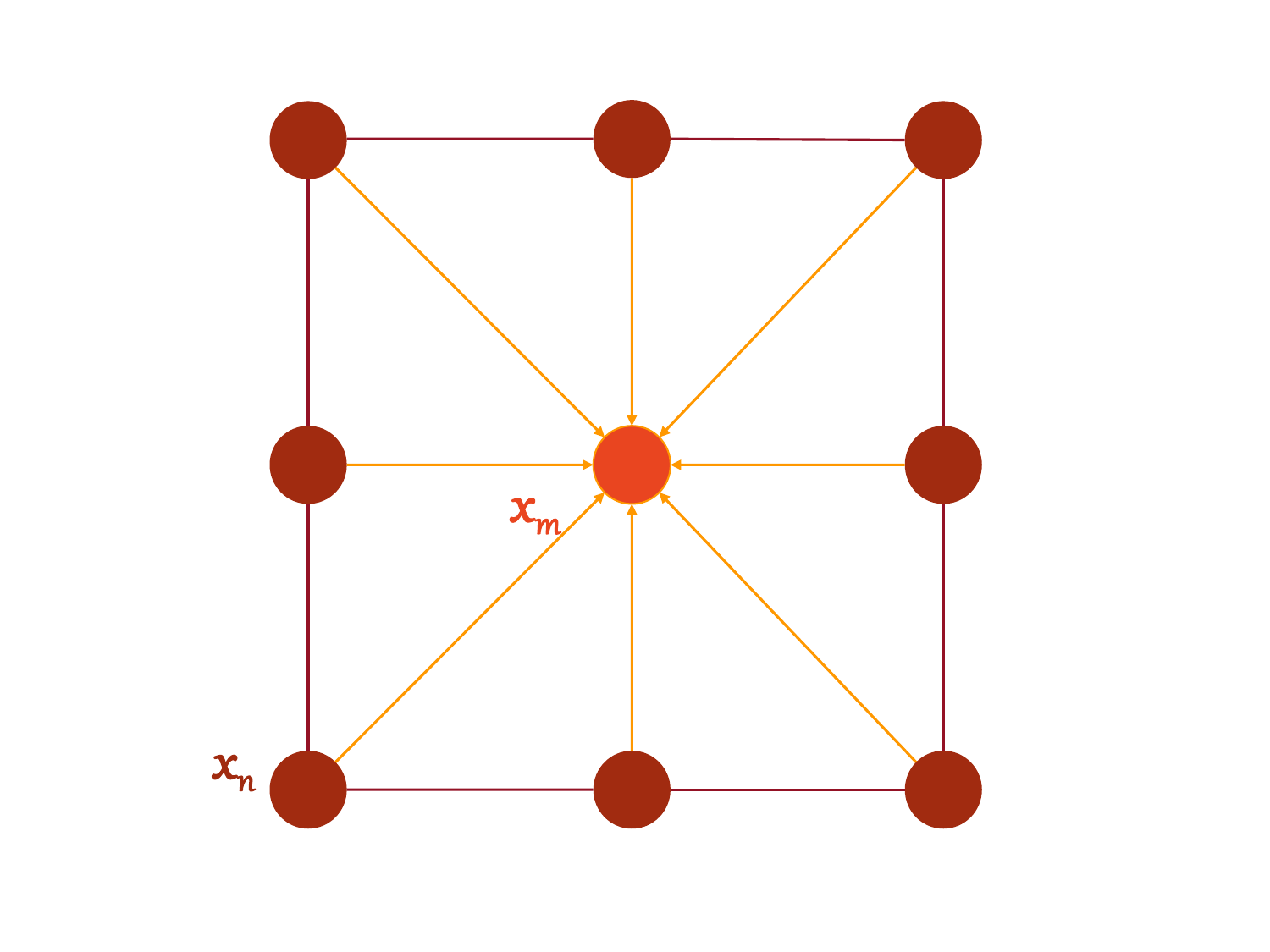}}
\caption{The special ‘lattice’ design  of patch-binary potential of AM module.  The probability of a specific patch (central patch in orange) is obtained by calculating the average of unary probabilities of eight neighbourhood patches.}
\label{fig:small2}
\end{figure}

The patch-binary potential $\psi_{(m,n)}({\mathrm x_{m}},{\mathrm x_{n}};\textbf{Y};w_{(m,n)};w_{E_{P}})$ of the Eq.~\ref{equ:1} illustrates the similarity of the pairwise neighboring patch sites $m$ and $n$ taking label $(\mathrm x_{m},\mathrm x_{n})=(c,c')$ given the data and weights, 
and it is defined as Eq.~\ref{equ:10}. 
\begin{equation}
\begin{aligned}
&\psi_{(m,n)}({\mathrm x_{m}},{\mathrm x_{n}};\textbf{Y};w_{(m,n)};w_{E_{P}})\propto{}\\
&\Big((p(\mathrm x_{(m,\alpha)}=c;\mathrm x_{(n,\alpha)}=c'|f_{(m,\alpha)}(Y),f_{(n,\alpha)}(Y)))^{w_{(m,n,\alpha)}}\\
&(p(\mathrm x_{(m,\beta)}=c;\mathrm x_{(n,\beta)}=c'|f_{(m,\beta)}(Y),f_{(n,\beta)}(Y)))^{w_{(m,n,\beta)}}\\
&(p(\mathrm x_{(m,\gamma)}=c;\mathrm x_{(n,\gamma)}=c'|f_{(m,\gamma)}(Y),
f_{(n,\gamma)}(Y)))^{w_{(m,n,\gamma)}}\Big)^{w_{E_{P}}},
\label{equ:10}
\end{aligned}
\end{equation}
where 
$\mathrm x_{n}=\{\mathrm x_{(n,\alpha)},\mathrm x_{(n,\beta)},\mathrm x_{(n,\gamma)}\}$ 
represents the patch labels and 
$ w_{(m,n)}=\{w_{(m,n,\alpha)},w_{(m,n,\beta)},w_{(m,n,\gamma)}\}$
 denotes the patch weights. 
The special design of eight neighbourhood structure is illustrated in Fig.~\ref{fig:small2}. The other operations are identical to the patch-binary potential. 

The core process of HCRF can be found in Algorithm~\ref{al1}.
\renewcommand{\algorithmicrequire}{\textbf{Input:}}
\renewcommand{\algorithmicensure}{\textbf{Output:}}
\begin{algorithm}[!htbp] 
\caption{HCRF} 
\label{al1}
\begin{algorithmic}[1] 
\Require  
The original image, $I$; 
The real label image, $L$;
\Ensure  
The image for segmentation result, $I_{seg}$; 
\State  
Put the original image $I$ into network and get $p(x_{i}=c|f_{i}(Y)$;
\For{pixel $i$ in the original image $I$}
\State Get $\varphi_{i}(x_{i};\textbf{Y};w_{V})$ defined as Eq.~\ref{equ:5};
\For{pixel $j$ in the neighbour nodes of pixel $i$}
\State Get $\psi_{(i,j)}(x_{i},x_{j};\textbf{Y};w_{E})$ defined as Eq.~\ref{equ:7};
\EndFor
\EndFor
\State Each pixel is taken as the center to get its corresponding patch; 
\State Put the original image $I$ into three networks and get $p(\mathrm x_{(m,\alpha)}=c|f_{(m,\alpha)}(Y))$, $p(\mathrm x_{(m,\beta)}=c|f_{(m,\beta)}(Y)$ and $p(\mathrm x_{(m,\gamma)}=c|f_{(m,\gamma)}(Y)$;
\For{patch $m$ in the original image $I$}
\State Get \quad
$\varphi_{m}({\mathrm x_{m}};\textbf{Y};w_{m};w_{V_{P}})$ defined as Eq.~\ref{equ:8};
\For{patch $n$ in the neighbour nodes of patch $m$}
\State Get \quad
$\psi_{(m,n)}({\mathrm x_{m}},{\mathrm x_{n}};\textbf{Y};w_{(m,n)};w_{E_{P}})$ defined as Eq.~\ref{equ:10};
\EndFor
\EndFor
\For{pixel $i$ in the original image $I$}
\State Get the corresponding patch $m$ of pixel $i$;
\State Get normalization factor $Z$ defined as Eq.~\ref{equ:2};
\State Get \quad
$
p(\textbf{X}|\textbf{Y})$ defined as Eq.~\ref{equ:1};
\State Get pixel-level classification result;
\EndFor
\State Get the image $I_{seg}$ for segmentation result;
\\ 
\Return $I_{seg}$; 
\end{algorithmic} 
\end{algorithm}

\subsection{IC Module}
\label{ss:method:IC}
Firstly, the abnormal images of the IC module in the training and validation dataset are sent to the trained HCRF model. 
The output of this step can be used to extract the diagnostic areas and direct the network's attention to the microscopic image classification. The next step is to threshold and mesh the output probability map. If the attention area occupies more than $50\%$ of the area of a $256 \times 256$ patch, this patch is chosen as the final attention patch (this parameter is obtained by traversing the proportion from $10\%$ to $90\%$ using the grid optimization method). Hence, the proposed HCRF-AM method can emphasize and give prominence to features that own higher discriminatory power.

Chemicals that are valuable for the diagnosis of gastric cancer, such as miRNA-215~\cite{deng2020mirna}, are also often expressed at higher levels in paracancerous tissue than in normal tissue~\cite{wang2018single}, indicating the significance of adjacent tissues for gastric cancer diagnosis. It indicates that only specific tumor areas for the networks conserved are insufficient. At the same time, we have tried to use only the extracted attention regions as the abnormal input image, but the results obtained prove that this is not feasible. Thus, all the images in the IC module dataset as well as the attention patches are used as input, which means the patches that are most likely to contain tumor areas are given more weight. Meanwhile, the neighboring patches of the attention patches will not be abandoned.

Transfer Learning (TL), one of the popular ML approaches, is developed to apply CNNs that are pre-trained on a large dataset with annotation of the same or different kinds of images (such as ImageNet) to accomplish multiple goals. TL concentrates on taking experience and knowledge from one problem and transferring it to a different but related field. In essence, it utilizes extra data to enable CNN to decode using features of past experience training, thus enabling CNN to have better generalization capability and higher efficiency~\cite{kamishima2009trbagg}.
Some popular DL networks such as the Inception series, VGG series, and ResNet series are taken into consideration in our work. The final choice is on the basis of numerous comparisons between their classification performance and the number of parameters. Finally, we select VGG-16 networks as our final classifier, whose parameters are pre-trained on the ImageNet dataset~\cite{deng2009imagenet}.
The pixel size of input images is $256 \times 256 \times 3$.

\subsection{Classification Probability-based Ensemble Learning (CPEL)}
As is mentioned above, the proposed HCRF-AM model is a patch-based network. Hence, it is necessary to propose an image-based classification method in order to calculate the probability that the input images contain tumors, or not. However, if we determine the possibility of the entire slice by merely calculating the average of all the scores in the probability graph, the results may be unreliable. To acquire accurate results in image-level from the patch-level output of CNNs, a CPEL algorithm is introduced~\cite{kittler1998combining}. The following equation can calculate the probability of each category
: 
\begin{equation}
\begin{aligned}
p(c_{j}\mid \textbf{Y}_{im})=\prod_{i=1}^{T}p(c_{j}\mid \textbf{Y}_{pa(i)})\propto \sum_{i=1}^{T}ln(p(c_{j}\mid \textbf{Y}_{pa(i)}))
\label{equ:3}
\end{aligned}
\end{equation}
Here,  $c_{j}$ denotes the image label ($c_{0}$ represents normal images and $c_{1}$ 
represents abnormal images). $\textbf{Y}_{im}$ is the input image with size of 2048 $\times$ 2048 pixels, 
and $\textbf{Y}_{pa}$  is the input patch with size of 256 $\times$ 256 pixels. T denotes the number of patches included in an input image. $p(c_{j}\mid \textbf{Y}_{im})$ represents the probability of an image labeled as normal or abnormal; Similarly, 
$p(c_{j}\mid \textbf{Y}_{pa})$ represents that of a patch. In addition, we take the logarithm of the probability in order to make sure that the numbers are within a certain range ($ln(\cdot)$ means the natural logarithm of a number). The final 
prediction is determined by the category which owns larger probability.

The whole process of our HCRF-AM framework is shown in Algorithm~\ref{al2}.
\begin{algorithm}[!htbp] 
\caption{HCRF-AM framework} 
\label{al2}
\begin{algorithmic}[1] 
\Require  
The image set for training and validation set of CNN with binary label, $\mathbb{I}$; 
The real label image set for abnormal images in $\mathbb{I}$, $\mathbb{L}$; 
The image set for test set of CNN,$\mathbb{I}_{test}$;
\Ensure  
The probability of an image labeled as normal or abnormal $p(c_{j}|\textbf{Y}_{im})$ of $\mathbb{I}_{test}$;
\State  
Divide $\mathbb{I}$ into abnormal image set $\mathbb{I}_{ab}$ and normal image set $\mathbb{I}_{nor}$ according to the binary label;
\For{image $I$ in $\mathbb{I}$}
\State Divide $I$ into patches and put them into CNN;
\If{$I \in\mathbb{I}_{ab}$}
\State Get real label image $L$ of $I$ from $\mathbb{L}$;
\State Put $I$ and $L$ into AM module and get segmentation result $I_{seg}$;
\State Divide $I_{seg}$ into patches and get patch set $\mathbb{P}_{seg}$;
\For{patch $P_{seg}$ in $\mathbb{P}_{seg}$}
\If{over 50\% pixels in $P_{seg}$ are segmented as abnormal regions}
\State $P_{seg}$ is chosen as attention region;
\State Put $P_{seg}$ into CNN;
\EndIf
\EndFor
\EndIf
\EndFor
\State Get trained CNN model;
\For{image $I_{test}$ in $\mathbb{I}_{test}$}
\State Put $I_{test}$ into CNN model and get patch-level classification result $p(c_{j}|\textbf{Y}_{pa(i)})$;
\State Get image-level classification result $p(c_{j}|\textbf{Y}_{im})$ defined as Eq.~\ref{equ:3};
\EndFor\\
\Return $p(c_{j}|\textbf{Y}_{im})$ of $\mathbb{I}_{test}$;
\end{algorithmic} 
\end{algorithm}

%% file: Experiment.tex
\section{Experiment}
\label{sec:Experiment}
\subsection{Experimental Settings}
\label{sec:Exp1}
\subsubsection{Dataset}
\begin{figure}[!htbp]
\centering
\centering{\includegraphics[width=1\columnwidth]{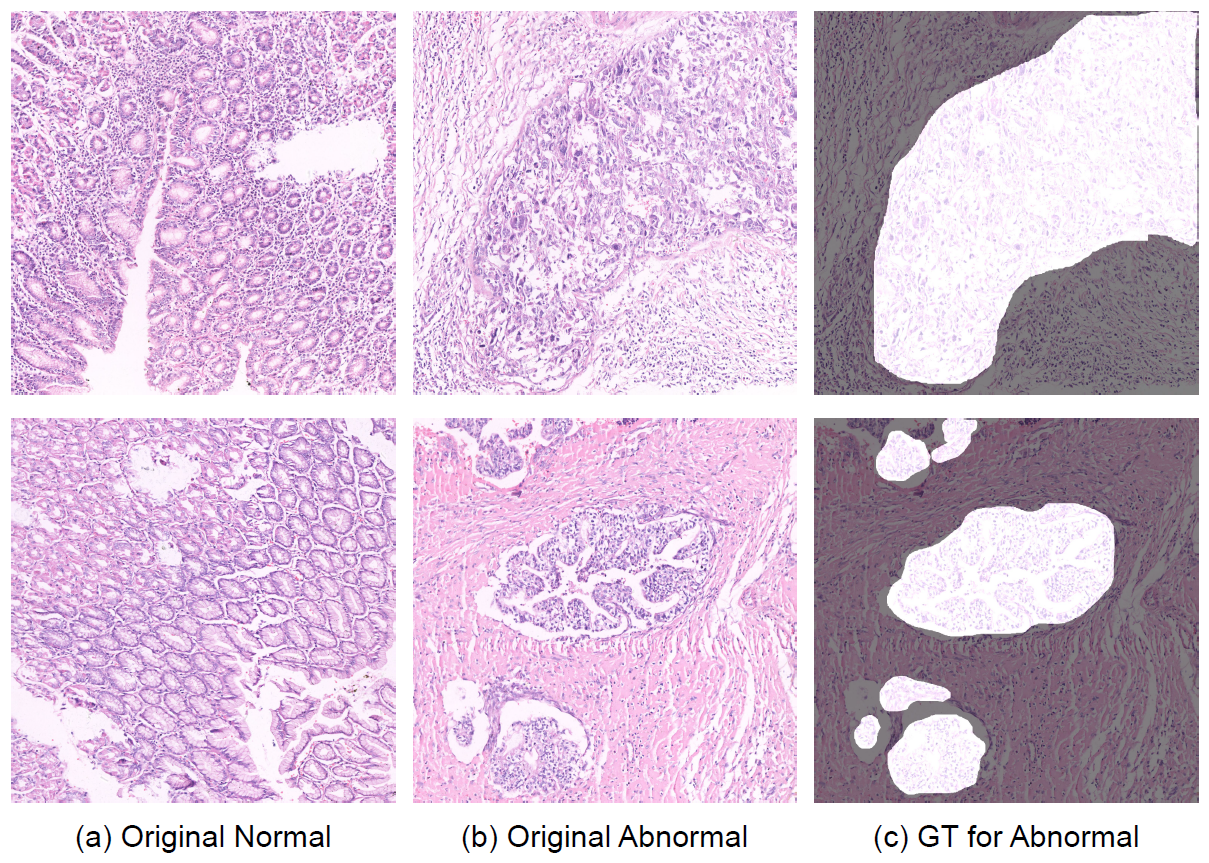}}
\caption{Example images of the data set used in the experiment. (a) shows the pathological microscopic images of normal gastric tissue. (b) shows pathological microscopic images of gastric tissue containing the cancerous area. (c) are the GT images one-to-one corresponding to (b). The brighter areas in the GT images are the cancerous regions while the darker areas are the parts that are not cancerous.}
\label{fig:dataset1}
\end{figure}

A public dataset containing 700 gastric histopathology images stained by H\&E is used in the research to build and evaluate the proposed HCRF-AM model~\cite{zhang2018pathological}, and some examples in the dataset are represented in 
Fig.~\ref{fig:dataset1}. 

These images are processed with H\&E, which is essential for identifying 
the diverse tissue types in histopathological images. As commonly used biological dye components, hematoxylin can dye the nucleus blue, while eosin can make the other parts outside the nucleus, including cytoplasm and extracellular matrix, present different shades of pink. \cite{fischer2008hematoxylin}. The images are magnified 20 times and 
most of the abnormal regions are marked by practical histopathologists. The image is in 
‘*.tiff’ or ‘*.png’ format and the image size is  $2048 \times 2048$ pixels. In the dataset, there 
are 140 normal images, 560 abnormal images attached with the corresponding GT images. 
In these normal images, cells are arranged regularly, the nucleo-cytoplasmic ratio is low, and a 
stable structure can be seen. By contrast, in the abnormal images, cancerous gastric tissue 
usually presents nuclear enlargement. Hyperchromasia without visible cell borders and prominent 
perinuclear vacuolization is also a typical feature~\cite{miettinen2006gastrointestinal}, \cite{Miettinen2003Gastrointestinal}. In the GT images, the cancerous areas in the tissue are marked by the pathologists.

\subsubsection{Experimental Design}
There are two main parts in the HCRF-AM model, AM module and IC module, so we distribute the images in the dataset according to our needs. The distribution is shown in Table~\ref{tab:data2}.
\begin{table}[!htbp]
\centering
\caption{The images allocation for AM module and IC module.}
\label{tab:data2}       
\begin{tabular}{lcc}
\toprule
\textbf{Image type} & \textbf{AM module} & \textbf{IC module}  \\
\midrule
Original normal images & 0 & 140 \\
Original abnormal images & 280 & 280 \\
\bottomrule
\end{tabular}
\end{table}

In the AM module, 280 abnormal images and corresponding GT images are used in the training of the HCRF model to obtain the attention areas, and the training set and the validation set are 140 images respectively, that is, the ratio is 1:1 (the detail information is in Sec.~\ref{ss:method:AM}). The AM module data setting is represented in Table~\ref{tab:data3}.
\begin{table}[!htbp]
\centering
\caption{The AM module data setting.}
\label{tab:data3}       
\begin{tabular}{lccc}
\toprule
\textbf{Image type} & \textbf{Train} & \textbf{Validation} & \textbf{Sum}  \\
\midrule
Original abnormal images & 140 & 140 & 280 \\
Augmented abnormal images & 53760 & 53760 & 107520\\
\bottomrule
\end{tabular}
\end{table}
 Before being sent into the model, the training and validation data sets are multiplied 6 times through rotation and mirror symmetry. In addition, since pathologists always observe on the patch scale when obtaining the characteristic information of cells in histopathological images, the original image and the GT image are cropped to a size of $256 \times 256$. After amplification, 53760 training images and the same number of validation images are obtained. 

In the IC module, 280 abnormal images remain and 140 normal images are applied in CNN classification part (the detail information is in Sec.~\ref{ss:method:IC}). The IC module data setting is represented in Table~\ref{tab:data4}.
\begin{table}[!htbp]
\centering
\caption{The IC module data setting.}
\label{tab:data4}       
\begin{tabular}{lcccc}
\toprule
\textbf{Image type} & \textbf{Train} & \textbf{Validation} & \textbf{Test} & \textbf{Sum}  \\
\midrule
Original normal images & 35 & 35 & 70 & 140 \\
Original abnormal images & 35 & 35 & 210 & 280 \\
Cropped normal images & 2240 & 2240 & -- & -- \\
Cropped abnormal images & 2240 & 2240 & -- & -- \\
\bottomrule
\end{tabular}
\end{table}
 Among them, 70 images from each class are randomly selected for training and validation sets, and the test set contains 70 normal images and 210 abnormal images. Similarly, we mesh these images into patches ($256 \times 256$ pixels). So, the initial dataset of the IC module comprises of 2240 training and 2240 validation images from each category. 
\subsubsection{Evaluation Method}
To evaluate our model, accuracy, sensitivity, specificity, precision and F1-score metrics are calculated to evaluate the classification result. The definition of the five indicators are given in Table~\ref{tab:data5}.

\begin{table}[!htbp]
\renewcommand\arraystretch{2}
\centering
\caption{The five evaluation indicators with definitions.}
\label{tab:data5}
\begin{tabular}{cccc}
\toprule
\textbf{Indicator} & \textbf{Definition} & \textbf{Indicator} & \textbf{Definition} \\
\midrule
\multirow{1}*{Accuracy} & $\dfrac{\mathrm{TP}+\mathrm{TN}}{\mathrm{TP}+\mathrm{FN}+\mathrm{TN}+\mathrm{FP}}$ &  {Sensitivity} & $ \dfrac{\mathrm{TP}}{\mathrm{TP}+\mathrm{FN}}$ \\
\multirow{1}*{Specificity}  &  $ \dfrac{\mathrm{TN}}{\mathrm{TN}+\mathrm{FP}}$  &  {Precision}  &  $ \dfrac{\mathrm{TP}}{\mathrm{TP}+\mathrm{FP}}$
\\
\multirow{1}*{F1-score}  &  $ \dfrac{\mathrm{2}\cdot\mathrm{Precision}\cdot\mathrm{Sensitivity}}{\mathrm{Precision}+\mathrm{Sensitivity}}$ &   &  
\\
\bottomrule
\end{tabular}
\end{table}

In this paper, positive samples are labeled as normal and negative samples are labeled as abnormal. In Table~\ref{tab:data5}, TP, namely True Positive, means that the normal image is identified as normal. TN, namely True Negative, indicating the abnormal image is identified as abnormal. FP, namely False Positive, that is abnormal image is identified as normal. FN, namely False Negative, that is normal image is identified as abnormal. Accuracy is defined as the proportion of correctly classified samples to the total sample. Sensitivity represents the proportion of correctly classified positive samples to all actual positive samples, and specificity means the proportion of all negative samples that are predicted correctly to all negative samples actually. Precision indicates the proportion of all samples classified as positive samples that are correctly judged. The F1 score takes into account the accuracy and recall of classification, and is defined as the harmonic average of the accuracy and recall of the model.

\subsection{Baseline Classifier Selection}
\label{sec:base}
\begin{figure}[!htbp]
\centering
\centering{\includegraphics[width=0.6\columnwidth]{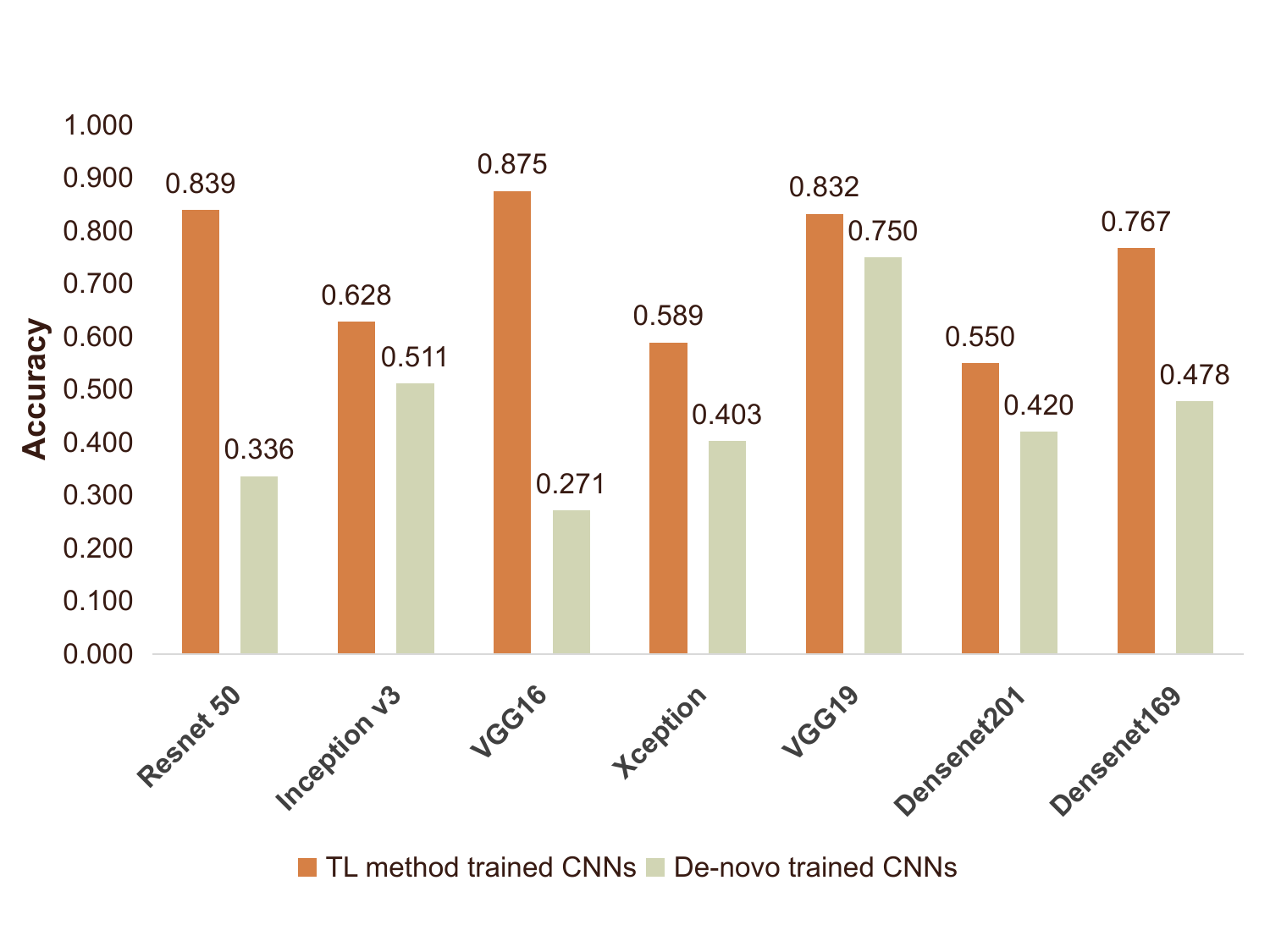}}
\caption{Comparison between image classification performance of different CNN-Based Classifiers on test set.}
\label{fig:histogram1}
\end{figure}

For baseline, we compare the performance between different CNN-based classifiers and evaluate the effect of Transfer Learning (TL) method on the initial dataset. We use the cropped  images in Table \ref{tab:data4} as the train and validation set to build the networks and the classification accuracy is obtained on the test set. The result is shown in Fig.~\ref{fig:histogram1}.

From Fig.~\ref{fig:histogram1}, it is observed that the VGG-16 TL method performs the best and achieves an accuracy of 0.875, followed by Resnet 50 and VGG-19 \cite{simonyan2014very} network. The Fig.~\ref{fig:histogram1} also proves that De-novo trained CNNs, which means training a convolutional neural network model from scratch, has worse performance in classification accuracy than TL algorithm. Therefore, the VGG-16 TL method is finally selected as the classifier in the baseline.

\subsection{Evaluation of AM Module}
\label{subsec:Evaluation1}
\begin{figure}[!htbp]
\centering
\centering{\includegraphics[width=0.9\columnwidth]{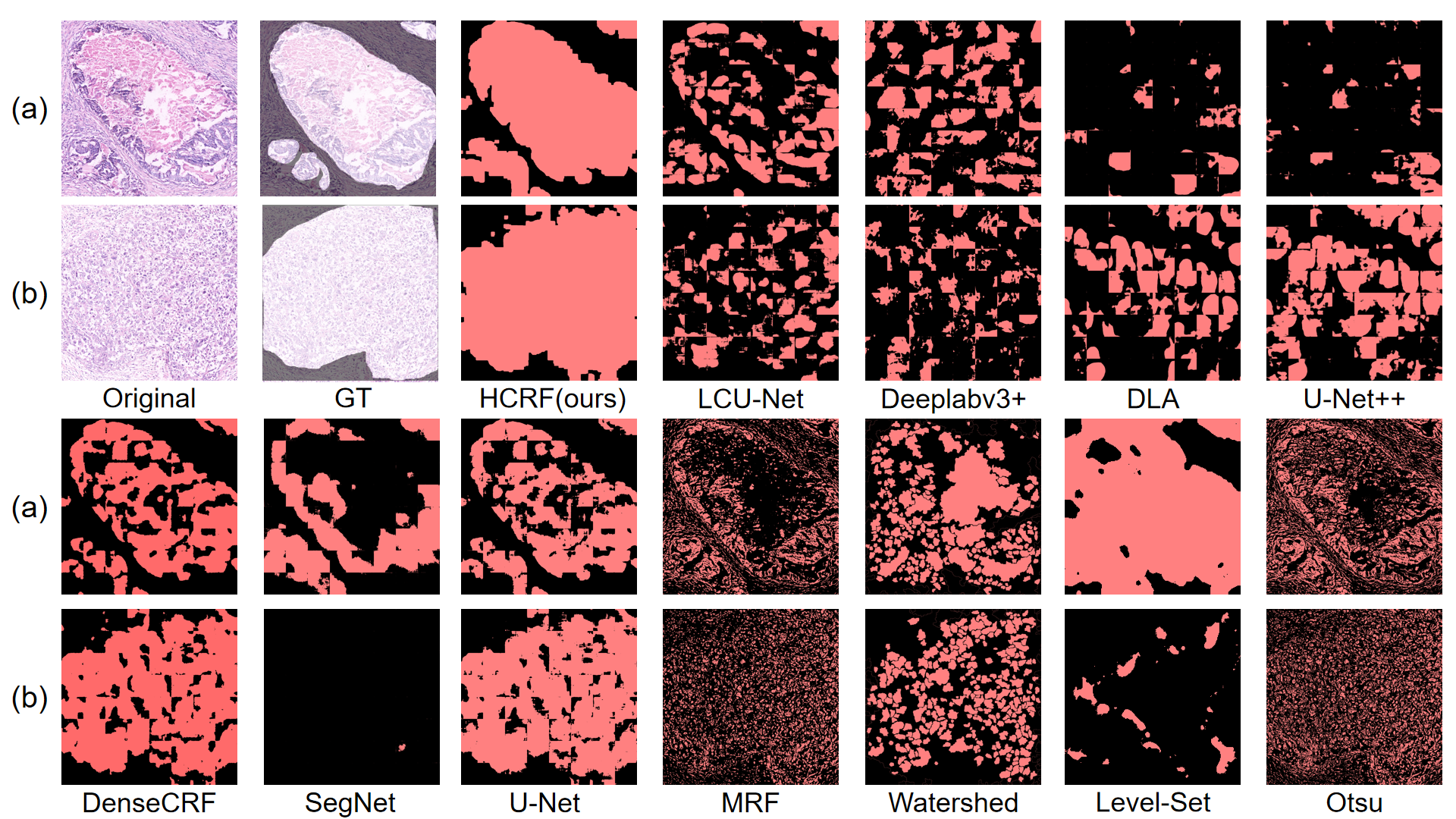}}
\caption{Comparison between HCRF and other attention area extracted methods on test set ((a) and (b) show 2 emblematic examples through various ways to extract the results of attention areas).}
\label{fig:seg}
\end{figure}

Based on our previous experiment~\cite{sun2020gastric}, we find that the HCRF model performs better than other advanced methods (LCU-Net~\cite{zhang2021lcu}, DLA(Unet+)~\cite{yu2018deep}, UNET++~\cite{zhou2018unet++} , Deeplabv3+~\cite{chen2018encoder}, DenseCRF~\cite{chen2017deeplab}, SegNet~\cite{badrinarayanan2017segnet}, and U-Net) and 4 classical methods (Level-Set~\cite{osher1988fronts}, Otsu thresholding~\cite{otsu1979threshold}, Watershed~\cite{vincent1991watersheds}, and MRF~\cite{li1994markov}) when segmenting interesting regions and objects. Fig.~\ref{fig:seg} shows the comparative analysis of the attention region extraction effect of our method and the above method on our data set. The advanced methods are all trained on the dataset in Table~\ref{tab:data3}.

Fig.~\ref{fig:seg} proves that our HCRF method can better extract the attention area, correctly mark more cancer areas, and retain less noise. The detailed information of evaluation index is shown in Table.~\ref{tab:AM}.
\begin{table}[!htbp]
\scriptsize
\centering
\caption{Comparison of image attention area extraction performance among HCRF model and some existing methods. The Dice is a measure of set similarity whose value range is [0,1]. IoU calculates the ratio of intersection and union of two sets. RVD is an asymmetric metric. The lower the RVD, the better the segmentation result. The numbers in bold in the table are the best for each indicator.}
\renewcommand\arraystretch{1.4}
\setlength{\tabcolsep}{3.5 pt}
    \centering
    \begin{tabular}{ccccccccc}
    \toprule
        \textbf{Architecture} & Year & Dice & IoU & Precision & Recall & Specificity & RVD & Accuracy \\ \midrule
        \textbf{Our HCRF} & 2021 & \textbf{0.4629} & \textbf{0.3259} & \textbf{0.4158} & 0.6559 & 0.8133 & \textbf{1.4135} & \textbf{0.7891} \\ 
        LCU-Net~\cite{zhang2021lcu} & 2021 & 0.4285 & 0.2891 & 0.2886 & 0.6167 & 0.8124 & 1.4967 & 0.7775 \\ 
        Deeplabv3+~\cite{chen2018encoder} & 2018 & 0.1597 & 0.0896 & 0.1583 & 0.2384 & 0.7622 & 1.9527 & 0.6778 \\ 
        DLA~\cite{yu2018deep} & 2018 & 0.1416 & 0.0795 & 0.1577 & 0.1981 & 0.8031 & 1.5678 & 0.7079 \\ 
        U-Net++~\cite{zhou2018unet++} & 2018 & 0.1564 & 0.0886 & 0.1542 & 0.2375 & 0.7569 & 1.9593 & 0.6761 \\
        DenseCRF~\cite{chen2017deeplab} & 2017 & 0.4578 & 0.3212 & 0.4047 & 0.6889 & 0.7812 & 1.6487 & 0.7702 \\ 
        SegNet~\cite{badrinarayanan2017segnet} & 2017 & 0.2008 & 0.1304 & 0.3885 & 0.3171 & \textbf{0.8412} & 2.0662 & 0.7531 \\ 
        U-Net~\cite{ronneberger2015u} & 2015 & 0.4557 & 0.3191 & 0.4004 & \textbf{0.6896} & 0.7795 & 1.6736 & 0.7684 \\ 
        MRF~\cite{li1994markov} & 1994 & 0.2396 & 0.1432 & 0.1839 & 0.4991 & 0.5336 & 4.5878 & 0.5441 \\ 
        Watershed~\cite{vincent1991watersheds} & 1991 & 0.2613 & 0.1585 & 0.2932 & 0.3541 & 0.7942 & 1.9434 & 0.7205 \\ 
        Level-Set~\cite{osher1988fronts} & 1988 & 0.2845 & 0.1920 & 0.2949 & 0.5202 & 0.7284 & 2.9296 & 0.6982 \\ 
       Otsu~\cite{otsu1979threshold} & 1979 & 0.2534 & 0.1505 & 0.2159 & 0.4277 & 0.7082 & 2.8859 & 0.6598 \\ \bottomrule
    \end{tabular}
    \label{tab:AM}
\end{table}

 The classical methods have similar results. The entire extracted region is scattered and abnormal areas cannot be separated. Some recent DL models are also be compared with our approach. The lower accuracy of LCU-Net may attribute to the decrease in the size of the parameter. Deeplabv3+ and DLA methods lack global information representation capabilities. To avoid the fusion of semantically dissimilar features of pure jump connections in U-Net, U-Net++ further strengthens the connections by nesting and jumping to reduce the the difference between encoding and decoding. Although effective performance has been achieved on some datasets, this method still cannot utilize enough information from multiple scales.  The SegNet requires large memory and not suitable for our data size.
 
In contrast, the HCRF model integrates the strengths of various models and takes the global information and contextual connection into consideration. Except for recall and specificity, our HCRF performs better than existing advanced methods in other indicators. The HCRF model has the best Accuracy and Recall, which shows its effectiveness in extracting attention regions' results and removing redundant information. 

In addition, based on the third-party experiments\cite{kurmi2020content}, the excellent performance of our HCRF model is also verified. In their experiments, the HCRF and other state-of-the-art methods (BFC \cite{zafari2015segmentation}, SAM \cite{wang2016semi}, FRFCM \cite{lei2018significantly}, MDRAN \cite{vu2019methods}, LVMAC \cite{peng2019local}, PABVS \cite{yu2020pyramid}, FCMRG \cite{sheela2020morphological}) are used for nuclei segmentation, and our HCRF model perform well, second only to the method proposed for their task in this experiment.

\subsection{Evaluation of HCRF-AM Model}

\begin{figure}[!htbp]
\centering
\centering{\includegraphics[width=0.9\columnwidth]{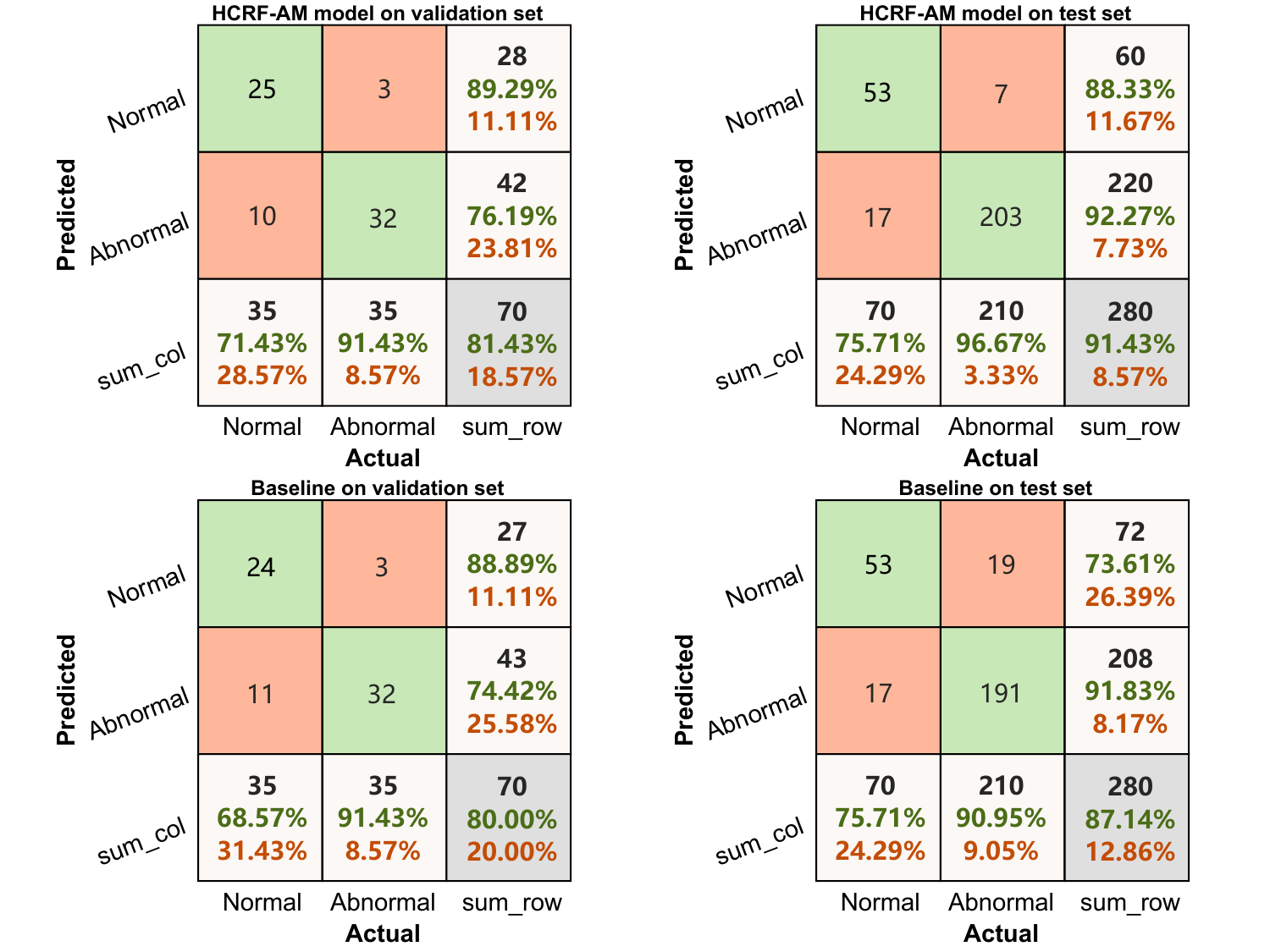}}
\caption{Confusion matrix of image classification on validation sets and test sets, presenting the classification results of baseline method and our HCRF-AM method, respectively.}
\label{fig:confu}
\end{figure}

\begin{figure}[!htbp]
\centering
\centering{\includegraphics[width=0.6\columnwidth]{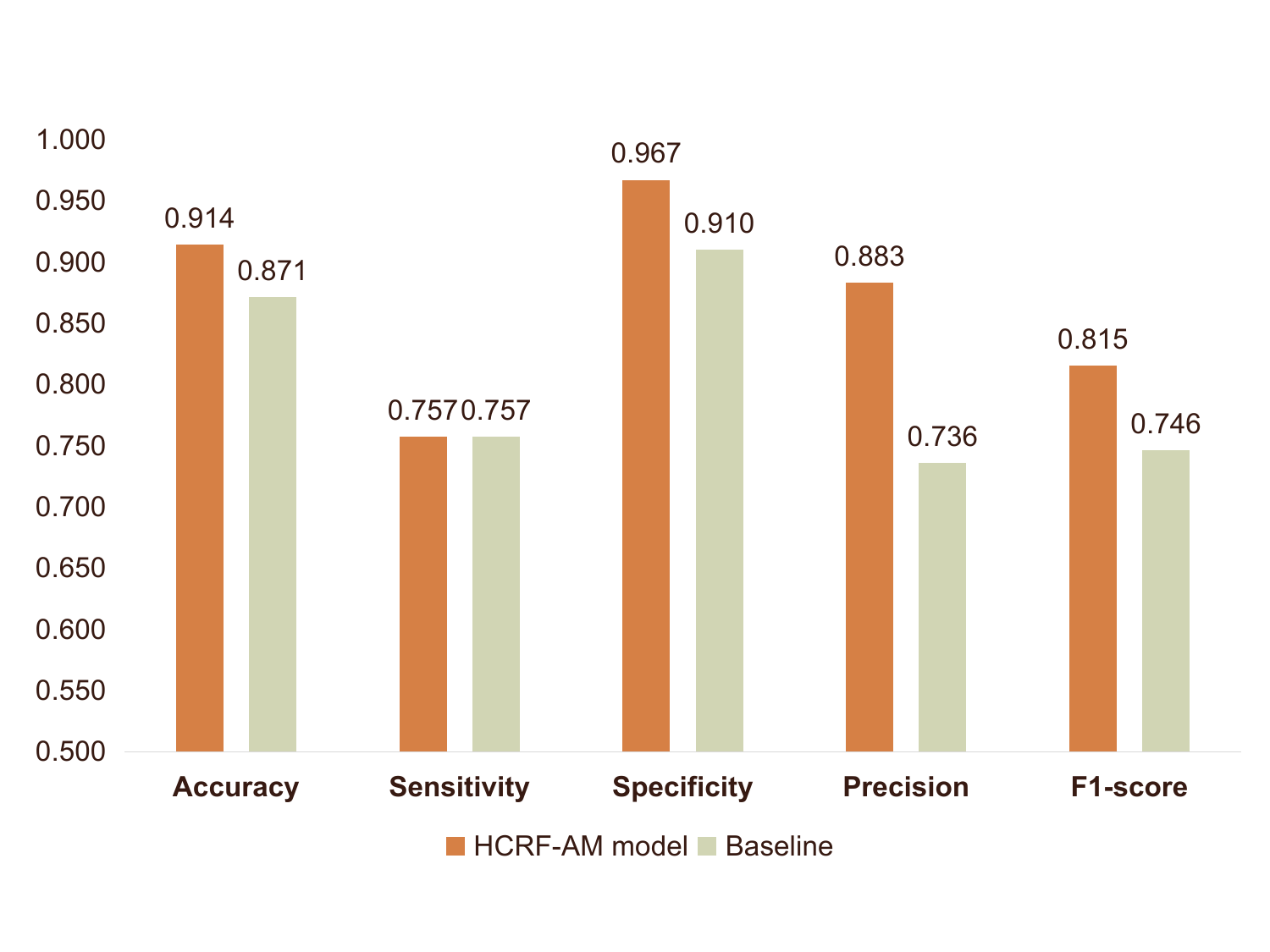}}
\caption{Comparison of HCRF-AM model and baseline image classification performance on the test set. }
\label{fig:his2}
\end{figure}

According to the experiment results in Sec.~\ref{sec:base}, we choose VGG-16 as our classifier in the IC module. Firstly, the cropped images in the training set and the validation set shown in Table.~\ref{tab:data4} as well as their attention areas are used to train the VGG-16 network with a TL algorithm. Due to the validation set, the CNN arguments are adjusted to restrain under-fitting or over-fitting during the training process. Then, $2048 \times 2048$ pixel images in the test sets are cropped into $256 \times 256$ pixel images and sent into the trained network to obtain the patch prediction probability. Thirdly, CPEL method is employed to acquire the final label of an image of $2048 \times 2048$ pixels. Finally, we evaluate the classification performance according to the true labels. In the first step, we use the grid optimization method to generate the best parameter of the proportion of attention area (the step size is 0.1). If the attention area takes up more than $50\%$ of the area of a patch, and this patch is chosen as the final attention patch.
A detailed view of comparison between baseline's and HCRF-AM model's classification accuracy and the corresponding confusion matrix is shown in Fig.~\ref{fig:confu} and Fig.~\ref{fig:his2}. For the VGG-16 network, the hyperparameter values are in Table.~\ref{tab:para}.
\begin{table}
    \centering
    \caption{The parameter settings for TL networks.}
    \begin{tabular}{lll}
    \toprule
        \textbf{Hyper-parameter} & VGG-16\\ \midrule 
        Initial input size & $256 \times 256 \times 3$ \\ 
        Initial learning rate & 0.0001 \\ 
        Batch-size & 64 \\ 
        Loss function & Categorical Cross-Entropy \\ 
        Opimizer & Adam~\cite{kingma2014adam} \\ 
        \bottomrule
    \end{tabular}
    \label{tab:para}
\end{table}

The confusion matrix in Fig.~\ref{fig:confu} shows our proposed method has better accuracy for the classification of normal and cancerous images. From Fig.~\ref{fig:his2}, we find that all evaluation indexes of our HCRF-AM model are about $1\%$ to $15\%$ higher than the baseline model.
The results show that although the test set has 4 times the number of images in the training and validation set, especially the abnormal images in the test set are 7 times that of the training and validation set, the HCRF-AM model we proposed still performs very high classification accuracy (especially the classification accuracy of abnormal images), representing good stability and robustness of the proposed method. Moreover, it has been verified the HCRF model attains better attention region extraction using GT images as standard in Sec.\ref{subsec:Evaluation1}. Table.~\ref{tab:result} shows a numerical comparison of the final classification results among our HCRF model and existing methods as attention extraction method on the test set. It proves that the HCRF method performs better on all indexes considering the final classification performance.
\begin{table}[!htbp]
    \scriptsize
    \caption{Numerical comparison of classification results between different attention extracted methods.}
    \label{tab:result}
    \renewcommand\arraystretch{1.4}
\setlength{\tabcolsep}{2.5 pt}
    \centering
    \begin{tabular}{ccccccc}
    \toprule
        \textbf{Architecture} & \textbf{Year} & \textbf{Accuracy} & \textbf{Sensitivity} & \textbf{Specificity} & \textbf{Precision} & \textbf{F1-score} \\ \midrule
        \textbf{Our HCRF} & 2021 & \textbf{0.914} & 0.757 & \textbf{0.967} & \textbf{0.883} & \textbf{0.815} \\ 
        LCU-Net~\cite{zhang2021lcu} & 2021 & 0.882 & 0.757 & 0.924 & 0.768 & 0.763 \\ 
        Deeplabv3+~\cite{chen2018encoder} & 2018 & 0.907 & \textbf{0.771} & 0.952 & 0.844 & 0.806 \\ 
        DLA~\cite{yu2018deep} & 2018 & 0.864 & 0.702 & 0.919 & 0.742 & 0.721 \\ 
        U-Net++~\cite{zhou2018unet++} & 2018 & 0.882 & 0.757 & 0.924 & 0.768 & 0.763 \\ 
        DenseCRF~\cite{chen2017deeplab} & 2017 & 0.893 & 0.686 & 0.962 & 0.857 & 0.762 \\ 
        SegNet~\cite{badrinarayanan2017segnet} & 2017 & 0.879 & 0.657 & 0.952 & 0.821 & 0.73 \\ 
        U-Net~\cite{ronneberger2015u} & 2015 & 0.893 & 0.729 & 0.948 & 0.823 & 0.773 \\ 
        MRF~\cite{li1994markov} & 1994 & 0.882 & 0.643 & 0.962 & 0.849 & 0.732 \\ 
        Watershed~\cite{vincent1991watersheds} & 1991 & 0.886 & 0.7 & 0.948 & 0.817 & 0.754 \\ 
        Level-Set~\cite{osher1988fronts} & 1988 & 0.896 & 0.714 & 0.957 & 0.847 & 0.775 \\ 
        Otsu~\cite{otsu1979threshold} & 1979 & 0.896 & 0.729 & 0.952 & 0.836 & 0.779 \\ \bottomrule
    \end{tabular}
\end{table}

\subsection{Comparison to Existing AM Methods}
\subsubsection{Existing Methods}
\label{subsec:ex-mthd}
Aiming to demonstrate the capabilities of our HCRF-AM model in GHIC tasks, we compared it with the recent DL network ViT~\cite{dosovitskiy2020AIIWW} and four latest AMs methods, including Squeeze-and-Excitation Networks (SENet)~\cite{hu2018squeeze}, Convolutional Block Attention Module (CBAM)~\cite{woo2018cbam}, Non-local neural networks (Non-local)~\cite{wang2018non} and Global Context Network (GCNet)~\cite{cao2019gcnet}. 
VGG-16 has a great number of parameters and it is hard to converse especially when integrated with other blocks~\cite{simonyan2014very}\cite{ioffe2015batch}. Based on the experiment constructed, we also find that training VGG-16 from scratch is a tricky problem. Meanwhile, the AMs nowadays have been extensively applied to Resnet and it is popular with the researchers~\cite{hammad2020resnet}\cite{roy2020attention}. Therefore, we combine these existing attention methods with Resnet in our contrast experiment in most cases.
 The following briefly introduces the main experimental parameters of these methods: (1) The original network of ViT is applied in this experiment. (2) SE blocks are integrated into a simple CNN with convolution kernel of $32\times 32, 64\times 64, 128\times 128, 256\times 256$ pixels. (3) CBAM is incorporated into Resnet v2 with 11 layers. (4) Nonlocal is applied to all residual blocks in Resnet with 34 layers. (5) GC blocks are integrated to Resnet v1 with 14 layers. They are all trained using the image data in Table~\ref{tab:data4} in the same way as when training our HCRF-AM model, and the input data size is $256\times 256$ pixels.

\subsubsection{Image Classification Result Comparison}
According to the experimental design in the Sec.~\ref{subsec:ex-mthd}, we obtained the experimental results in the Table~\ref{tab:compare}.

\begin{table}[!htbp]
\caption{Comparison of classification effect between HCRF-AM model and other methods on gastric histopathology image test set.}
    \label{tab:compare}
    \centering
    \begin{tabular}{ccccc}
    \toprule
        \textbf{Method} &\textbf{Year} & \textbf{Accuracy} & \textbf{Sensitivity} & \textbf{Specificity }\\ \midrule
       \textbf{HCRF-AM(Our method)}  & 2021 & \textbf{0.914} & 0.757 & \textbf{0.967} \\
        ViT \cite{dosovitskiy2020AIIWW} & 2020 & 0.802 & \textbf{0.881}& 0.791 \\ 
        GCNet+Resnet \cite{cao2019gcnet} & 2019 & 0.741 & 0.557 & 0.811 \\ 
        SENet+CNN \cite{hu2018squeeze} & 2018 & 0.754 & 0.429 & 0.862 \\ 
        CBAM+Resnet \cite{woo2018cbam} & 2018 & 0.393 & 0.843 & 0.243 \\ 
        Non-local+Resnet \cite{wang2018non} & 2018 & 0.725 & 0.571 & 0.776\\
         \bottomrule
    \end{tabular}
\end{table}

Table~\ref{tab:compare} proves that: (1) Compared with five advance existing methods, except sensitivity, the HCRF-AM constructed in this article has the highest value in other indicators. The overall accuracy of most methods is around 70\%, apparently lower than that of ours. (2) The sensitivity of HCRF-AM is the second best only after ViT model, and the other two indicators of ViT are far lower than us. And in practical diagnosis, the specificity, which reflects the abnormal case of correct judgement, is of particular importance. (3) The sensitivity and specificity of SE blocks and GC blocks vary widely, whose differences are around 30\%. This suggests that their prediction strategy is out of balance (see further discussion in Sec.~\ref{subsec:method:exist}).

\subsection{Computational Time}
A workstation with Intel$^{ \circledR }$ Core$^{TM}$ i7-8700k 
CPU 3.20 GHz, 32GB RAM and GeForce RTX 2080 8 GB is used to complete the experiment. The acquisition of the final model requires the completion of the training of the two modules, the AM module and IC module, taking about 50 h for training 280 images ($2048 \times 2048$ pixels) and 1431 s for training 140 images ($2048 \times 2048$ pixels), respectively. The AM module and IC module take 
about 50 h and 1431 s for training, respectively. It takes 0.5 seconds on average to classify an image ($2048 \times 2048$ pixels) on the test set through the HCRF-AM model. From the data of calculation time above, it can be seen that our model needs time in the training phase, but the classification in the testing phase after the model is prepared is very efficient, which shows that the model is suitable for real-time clinical diagnosis.

%% file: discussion.tex
\section{Discussion}
\label{sec:Discussion}

\subsection{Analysis of the Dataset}
The experiment results in Sec.~\ref{subsec:Evaluation1} suggest there is still a lot of room for improvement of AM module in choosing attention areas. Hence, we invite our cooperative histopathologists to analyze this open-source dataset and determine the reason for mis-segmentation.

\begin{figure}[!htbp]
\centering
\centering{\includegraphics[width=0.9\columnwidth]{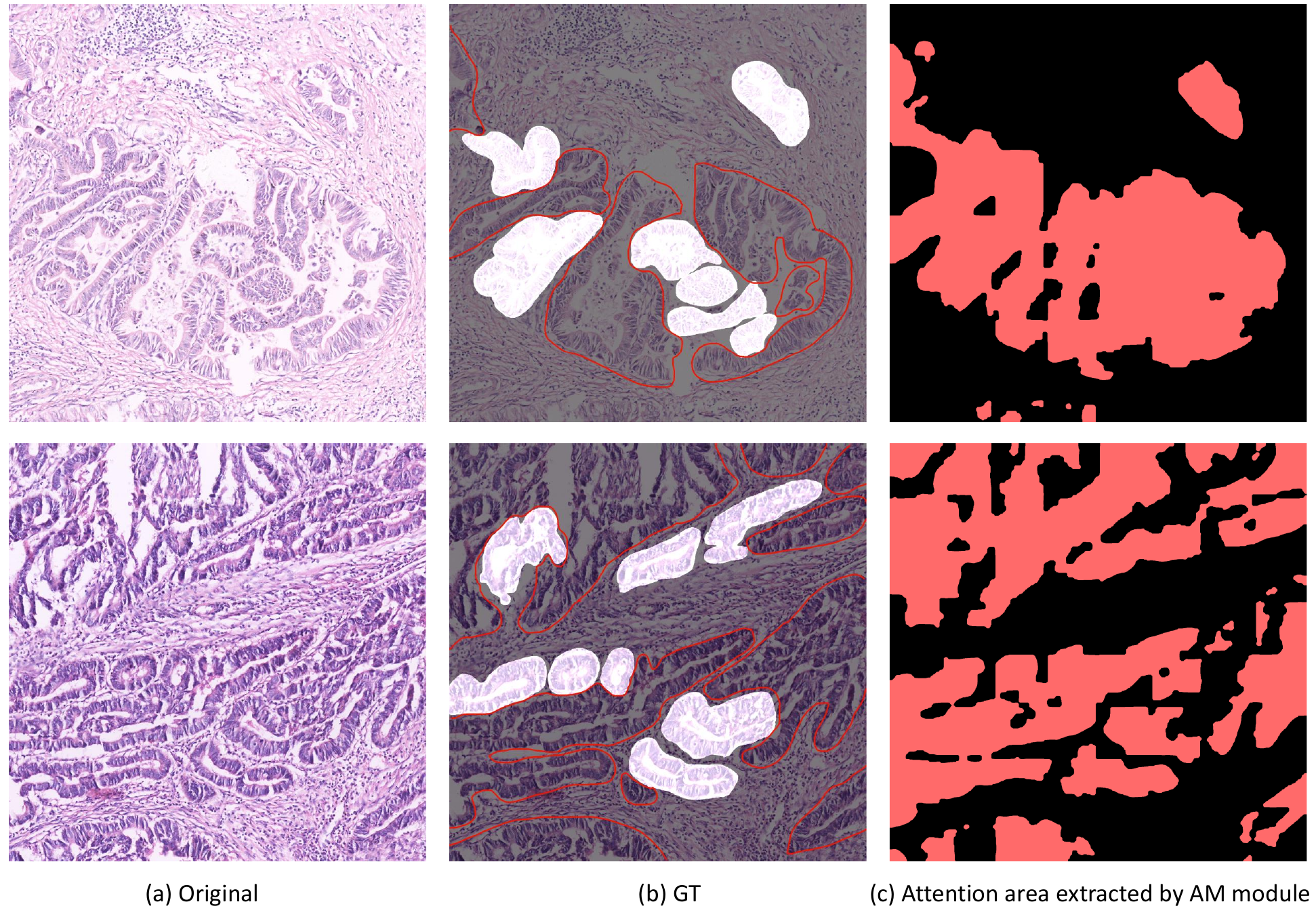}}
\caption{Typical examples of some images in our dataset for analysis. (a) denotes the original images. (b) presents the GT images. The cancerous areas relabeled by our collaborating histopathologists are shown as areas surrounded by red curves. The red regions of (c) shows the attention extraction results by the AM module. }
\label{fig:dataerror}
\end{figure}

According to Fig.~\ref{fig:dataerror} and our cooperative histopathologists' medical knowledge, pathologists are often incomplete in marking the abnormal areas of gastric histopathological images. This omission is especially obvious when there are many cancerous areas in an image. Doctors often only mark the most representative and risky cancerous areas, which causes some actual cancerous areas to be ignored and marked as normal area, as shown in Fig.~\ref{fig:dataerror} (b). 
Fig.~\ref{fig:dataerror} (b) and (c) indicate our HCRF model in AM module may extract the attention areas correctly, however the GT image marked by the doctor may ignore part of the area due to the coarse labels. We confirm the above situation to the histopathologists we worked with, and ask them to remark the cancerous areas in the original data set with red curves. Comparing the areas we ask pathologists to relabel with the marked areas in the public data set, we can see that the GT images provided by the public data set ignore a large number of cancerous areas when labeling. Fig.~\ref{fig:dataerror} shows that the results of our attention region extraction and the relabeled image cancerous regions have a higher degree of overlap, and the problem of GT images in the dataset may be the reason for the insufficient IoU and RVD indicators.
\subsection{Mis-classification Analysis}
\begin{figure}[!htbp]
\centering
\centering{\includegraphics[width=1\columnwidth]{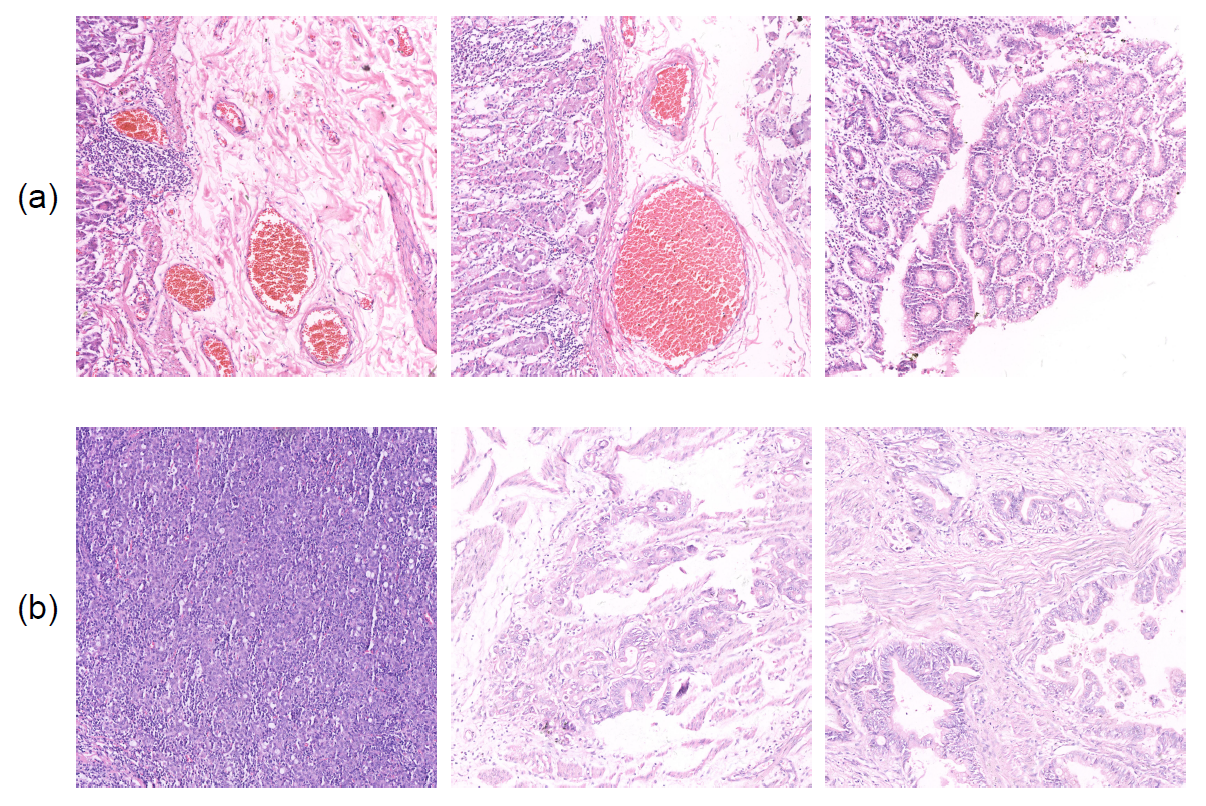}}
\caption{Examples of mis-classification. The row (a) presents the normal cases diagnosed as abnormal (FN). The row (b) presents the abnormal cases diagnosed as normal (FP).}
\label{fig:dataset}
\end{figure}

To analyze the causes of mis-classification, some examples is given in Fig.~\ref{fig:dataset}. For FN samples in Fig.~\ref{fig:dataset}(a), some larger bleeding spots can be found in some normal samples, leading to misdiagnosis. Some images have many bright areas in the field of view, which may be caused by being at the edge of the whole slice, and these bright areas cannot provide information effectively. For FP samples in Fig.~\ref{fig:dataset}(b), the cancer areas in some images for abnormal samples are small and scattered, making them insufficiently noticed in classification. Simultaneously, in some samples, the staining of the two stains is not uniform and sufficient. In some images, diseased areas appear atypical, which increases the difficulty of classification. 

\subsection{Analysis of the Existing Attention Mechanisms}
\label{subsec:method:exist}
Recently, AMs have drawn significant attention from scholars and are widely used when solving problems in different fields. For example, the non-local network is proposed to model the long-range dependencies using one layer via a self-AM~\cite{wang2018non}. However, with the increasing area of the receptive field, the computation costs become more extensive at the same time. These AMs, which have a large memory requirement, are not suitable in GHIC tasks because the size of histopathology images are always $2048 \times 2048$ pixels or even larger. Unlike the natural image, resizing is fatal to histopathological images, which may risk losing many essential details and textures. Another limitation exists in the lack of pre-trained models when we utilize TL methods in these existing AMs. Training-from-scratch makes the neural network tricky to convergence, let alone achieve good performance~\cite{mishkin2015all}. There are some methods proposed to solve this problem, such as SE block. The feature importance induced by SE blocks used for feature recalibration highlights useful features and restrain redundancy, which causes a decrease in computation complexity. Furthermore, it is simple to implement the SE block, which has strong compatibility with the current high-performance DL model~\cite{hu2018squeeze}. As for the network ViT that is compared in Sec.~\ref{subsec:ex-mthd}, it has a strong ability to describe global information. However, the demand for training data is often higher than the commonly used CNN. Therefore, in the GHIC task, the performance is not so satisfactory.

%% file: conclusion.tex
\section{Conclusion and Future Work}
\label{sec:Conclusion}
In this paper, we develop a novel approach for GHIC using an HCRF-based AM. Through experiments, we choose high-performance methods and networks in the AM and IC modules of the HCRF-AM model. Our HCRF method surpasses the progressive attention area extraction methods in the evaluation process, showing the robustness and potential of the proposed method. Finally, our method attains a classification accuracy of $91.4\%$ and a specificity of $96.7\%$ on the testing images. Our proposed method has been compared with some existing popular AMs methods that use the same dataset to verify the performance. In summary, the HCRF-AM model has the feasibility of being employed for computer-assisted diagnosis of early gastric cancer, which may help improve the work efficiency of pathologists. In the discussion part, the possible causes of misclassification in the experiment are analyzed, which provides a way to improve the performance of our model.

Though our method provides satisfactory performance, there are a few limitations. First, our proposed HCRF model in the AM module only considers information on a single scale, which degrades model performance. Moreover, our model can be further improved by the technique shown in~\cite{Zormpas2019Superpixel}, where large-scale tissue and cross-spatial scale information are introduced into cell classification tasks. Second, we have investigated four kinds of DL models, using TL methods and integrating the AM into them. In the future, we can investigate other DL models and compare their results for higher classification accuracy. Finally, our AM is a weakly supervised system at present. Hence, the unsupervised learning method~\cite{dosovitskiy2020image} may be of certain reference significance to ours, which applies a pure transformer directly to sequences of image patches and performs well on natural image classification tasks. Simultaneously, by modifying the network, adding weights to the network to find areas with high weights will also be our future research direction.

%% file: ack.tex
\begin{acknowledgements}
This study was supported by the National Natural
Science Foundation of China (grant No. 61806047). 
We thank Miss Xiran Wu, due to her contribution is considered as 
important as the first author in this paper. 
We also thank Miss Zixian Li and Mr. Guoxian Li for their important discussion.
\end{acknowledgements}